\documentclass{article}

\usepackage{microtype}
\usepackage{graphicx}
\usepackage{subfigure}
\usepackage{booktabs} 
\usepackage{multirow}
\usepackage{bbm}
\usepackage{amsmath}
\usepackage{amssymb}
\usepackage{xcolor}

\usepackage{hyperref}

\usepackage{enumitem}
\setlist{nolistsep}
\usepackage{amsthm}
\theoremstyle{definition}
\newtheorem{exmp}{Example}[section]

\usepackage[show]{chato-notes}
\usepackage{cleveref}
\crefname{figure}{Fig.}{Figs.}
\Crefname{figure}{Figure}{Figures}
\Crefname{section}{Section}{Sections}
\crefname{section}{Sect.}{Sect.}
\crefname{subsection}{Sect.}{Sect.}
\Crefname{subsection}{Section}{Sections}
\crefname{subsubsection}{Sect.}{Sect.}
\Crefname{subsubsection}{Section}{Sections}
\crefname{table}{Table}{Tables}
\Crefname{table}{Table}{Tables}
\crefname{exmp}{Example}{Examples}
\Crefname{exmp}{Example}{Examples}

\newcommand{\sys}{TAGLETS\xspace}
\newcommand{\scads}{SCADS\xspace}

\newcommand{\model}{h}
\newcommand{\xtarget}{\mathcal{X}_T}
\newcommand{\ytarget}{\mathcal{Y}_T}
\newcommand{\xaux}{\mathcal{X}_S}
\newcommand{\yaux}{\mathcal{Y}_S}

\newcommand{\labsetname}{\mathcal{X}}
\newcommand{\labset}{\labsetname=\{(x, y) \in (\xtarget, \ytarget)\}}
\newcommand{\unlabsetname}{\mathcal{U}}
\newcommand{\unlabset}{\unlabsetname = \{x \in \xtarget\}}
\newcommand{\auxsetname}{\mathcal{A}}
\newcommand{\auxset}{\auxsetname = \{(x, y) \in (\xaux, \yaux)\}}
\newcommand{\selectedauxsetname}{\mathcal{R}}

\newcommand{\graph}{G}
\newcommand{\nodes}{Q}
\newcommand{\edges}{E}
\newcommand{\concept}{q}
\newcommand{\setconcept}{Q}
\newcommand{\class}{c}

\newcommand{\scadsemb}[1]{\hat{e}_{#1}}
\newcommand{\wordemb}[1]{e_{#1}}
\newcommand{\neigh}{\mathcal{N}}
\newcommand{\objemb}[1]{\Psi(#1)}
\newcommand{\numshots}{K}
\newcommand{\numrelated}{N}

\newcommand{\module}{m}

\newcommand{\taglet}[1]{t_#1}
\newcommand{\tagletcollection}{\mathcal{T}}
\newcommand{\encoder}{\phi}
\newcommand{\weightsencoder}{\theta}
\newcommand{\loss}[1]{\mathcal{L}_{#1}}

\newcommand{\vote}{V}

\usepackage[accepted]{mlsys2022}

\mlsystitlerunning{TAGLETS: A System for Automatic Semi-Supervised Learning with Auxiliary Data}

\begin{document}

\twocolumn[
\mlsystitle{TAGLETS: A System for Automatic \\ Semi-Supervised Learning with Auxiliary Data}



\mlsyssetsymbol{equal}{*}

\begin{mlsysauthorlist}
\mlsysauthor{Wasu Piriyakulkij}{brown}
\mlsysauthor{Cristina Menghini}{brown}
\mlsysauthor{Ross Briden}{brown}
\mlsysauthor{Nihal V. Nayak}{brown}
\mlsysauthor{Jeffrey Zhu}{brown}
\mlsysauthor{Elaheh Raisi}{linkedin}
\mlsysauthor{Stephen H. Bach}{brown}
\end{mlsysauthorlist}

\mlsysaffiliation{brown}{Department of Computer Science, Brown University, Providence, RI, USA}
\mlsysaffiliation{linkedin}{LinkedIn, Sunnyvale, CA, USA}

\mlsyscorrespondingauthor{Wasu Piriyakulkij}{wasu\_piriyakulkij@brown.edu}

\mlsyskeywords{Transfer learning, few-shot learning, semi-supervised learning}

\vskip 0.3in

\begin{abstract}
Machine learning practitioners often have access to a spectrum of data: labeled data for the target task (which is often limited), unlabeled data, and \emph{auxiliary data}, the many available labeled datasets for other tasks.
We describe \sys, a system built to study techniques for automatically exploiting all three types of data and creating high-quality, servable classifiers.
The key components of \sys are: (1) auxiliary data organized according to a knowledge graph, (2) modules encapsulating different methods for exploiting auxiliary and unlabeled data, and (3) a distillation stage in which the ensembled modules are combined into a servable model.
We compare \sys with state-of-the-art transfer learning and semi-supervised learning methods on four image classification tasks.
Our study covers a range of settings, varying the amount of labeled data and the semantic relatedness of the auxiliary data to the target task.
We find that the intelligent incorporation of auxiliary and unlabeled data into multiple learning techniques enables \sys to match---and most often significantly surpass---these alternatives.
\sys is available as an open-source system at \url{github.com/BatsResearch/taglets}.

\end{abstract}
]



\printAffiliationsAndNotice{}

\section{Introduction}\label{sec:intro}

The bottleneck of labeled training data in machine learning has driven research on exploiting alternative, more abundant sources of data such as unlabeled data and \emph{auxiliary data}, i.e., labeled data for possibly related tasks other than the target task.
As unlabeled and auxiliary data becomes abundant, labeled training data for a given task has come to represent only a small fraction of the data available to practitioners.
As we show, there are underexplored opportunities for taking advantage of all these resources in an integrated, data-centric manner.
In this work, we propose a system for automatically exploiting the many annotated datasets either publicly accessible or internally available to organizations, as well as any available unlabeled data, for novel 
tasks.

\begin{figure}[t]
\centering
\includegraphics[width=0.45\textwidth]{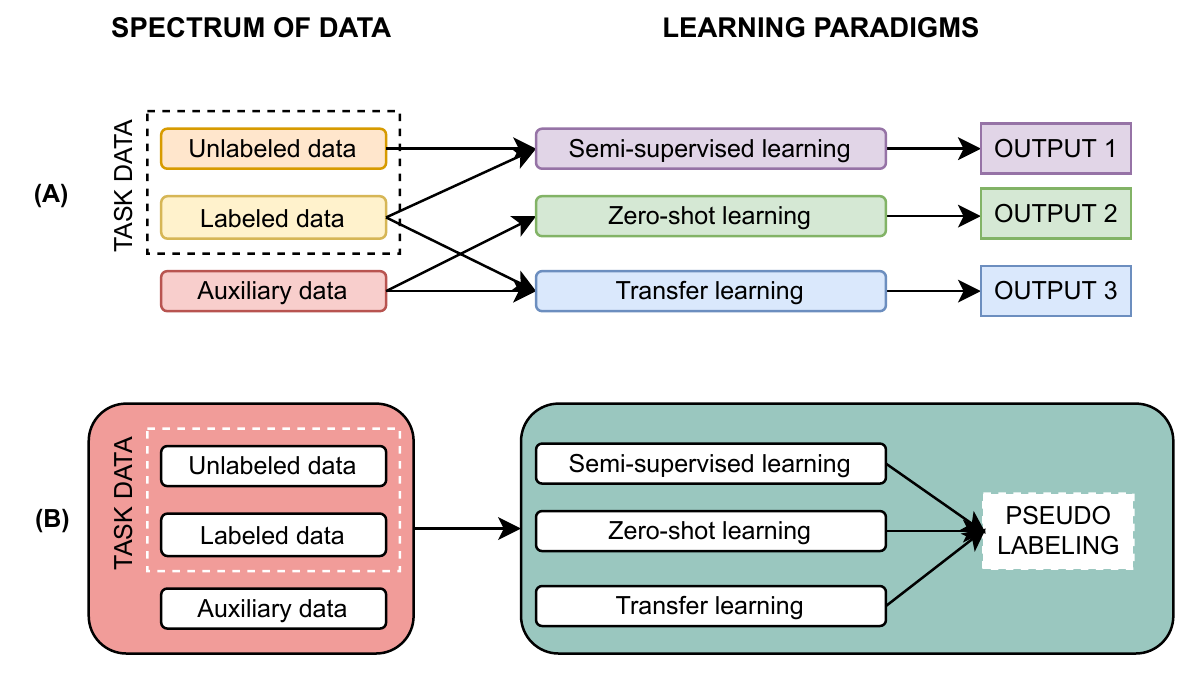}
\caption{\textbf{(A)} \emph{Common practice}. On the left, three categories of data separately passed as input to methods belonging to diverse learning approaches. Each method returns a distinct output.\\ \textbf{(B)} \emph{The envisioned approach}. Methods can be trained on all available data. The predictions of such methods are then combined to produce the intermediate pseudo labels for training an end model.
}
\label{fig:data}
\end{figure}

In current practice, optimally taking advantage of the full spectrum of available data is not straightforward.
Distinct areas of machine learning have grown up around scenarios emphasizing access to different types of data: transfer learning~\citep{pan:tkde10,yosinski:neurips14,weiss:bigdata16,devlin:naacl19} for exploiting auxiliary data and semi-supervised learning~\citep{chapelle:book06,sohn:fixmatch,chen:simclrbig,pham:metapseudo} for a mix of labeled and unlabeled data.
Transfer learning has become the default option in machine learning, and it offers a simple recipe: pre-train a model on a large labeled dataset and then fine-tune it on the target task.
This approach is useful for learning widely reusable features of a domain, but is also limiting because it does not take into account the fine-grained similarities between the target task and subsets of the available auxiliary data.
Further, it does not take advantage of complementary resources and techniques, such as semi-supervised learning and zero-shot learning~\citep{xian:pami18}.

Automatically exploiting auxiliary data, unlabeled data, and limited labeled data for the target task simultaneously requires addressing three main challenges:
\begin{enumerate}
    \item \textbf{Relevant auxiliary data is a needle in a haystack.}
    Often a huge number of auxiliary datasets are available, but likely only a very small subset is significantly related to any given target task.
    Further, related examples in auxiliary datasets can be useful for many distinct reasons.
    In the object classification setting, related datasets might contain examples of visually similar objects, functionally similar objects, objects likely to co-occur in similar settings, and more.
    How can we automatically find the relevant data quickly?
    \item \textbf{Best practices vary by amount of training data.} The most accurate methods for learning with limited labeled data vary according to several factors, including how much labeled data is available and how related auxiliary data is to the target task.
    How can we automatically be robust to these variations?
    \item \textbf{Pipelines are difficult to serve in production.}
    In low-latency applications with service-level agreements, it is necessary to make predictions in a fixed amount of time.
    This requirement limits the complexity of prediction pipelines that are practical.
    How can we distill all the information in labeled training data, unlabeled data, and auxiliary data into a single, servable model?
\end{enumerate}

In this work, we address these challenges with \sys, a modular end-to-end machine learning system for automatic semi-supervised learning with auxiliary data.

\sys automatically selects relevant subsets of auxiliary data and makes them available to semi-supervised, transfer, and zero-shot learning modules.
The modules create an intermediate pseudo labeling, and then \sys distills them into a final, servable model~(\Cref{fig:data}).
The system is designed to work with limited labeled data, instead relying on the auxiliary annotated datasets publicly available and internal to organizations.
Additionally, unlabeled data from the task distribution is used to further boost performance.

\sys is targeted at scenarios in which non-experts in machine learning need to create classifiers that are deployable at scale.
They have access to abundant auxiliary data because much of it is publicly available, and they or their organization might have collected more internally.
Use cases include creating classifiers for data science projects and rapid prototyping.
We have distilled principles that guided the design of \sys from our experience working with mixtures of labeled, unlabeled, and auxiliary data.
\begin{enumerate}
    \item \textbf{Select the best auxiliary data available.} In some cases, a target task might consist of classes that overlap substantially with those in auxiliary data.
    In others, the auxiliary might be only more loosely related.
    The system should allow models to transfer knowledge from auxiliary data in both cases.
    \item \textbf{Learn from limited target data.} Different scenarios have access to different amounts of labeled data.
    The system should be effective when learning from one training example per target class to dozens.
    \item \textbf{Automatically distill to a servable model.} A user should be able to create a model by providing a minimal semantic description of their target task, and any available labeled and unlabeled data.
    The system should then automatically produce a high-quality model small enough to be served in production.
\end{enumerate}

We designed \sys to address the above challenges and principles.
The key contributions of our work are:

\textbf{Structured collection of datasets.} We introduce the concept of a structured collection of annotated
datasets (\scads).
A \scads gathers all available labeled datasets and connects concepts described in those datasets via a knowledge graph. 
This repository of linked data is where \sys automatically sources the auxiliary data.
\scads is described in~\Cref{sec:scads}.

\textbf{Ensemble of methods exploiting auxiliary data.}
We propose ensembling methods that can exploit a SCADS for robust predictive accuracy.
We show how to adapt existing transfer and semi-supervised learning methods to exploit a SCADS in a modular framework.
Zero-shot learning based on knowledge graphs further boosts accuracy.
We show that ensembling these methods is key to robust accuracy across a range of scenarios covering different amounts of labeled target data and relatedness of the data in the SCADS to the target task.
These methods are described in~\Cref{sec:modules}.

\textbf{End-to-end system.} \sys is the first automatic end-to-end machine learning system that simultaneously exploits ecosystems of auxiliary data and models.
It achieves state-of-the-art performances on practically useful tasks with limited labeled data.
We release it as an open-source package at \url{github.com/BatsResearch/taglets}.

In this paper, we instantiate \sys for object classification tasks.
We use ConceptNet~\citep{speer:conceptnet}, a common sense knowledge graph, and ImageNet 21k~\citep{russakovsky:imagenet} to create a \scads.
We evaluate \sys by comparing with state-of-the-art transfer and semi-supervised learning methods on four practical object classification tasks.
Our study includes experiments with a range of labeled data for target tasks (from 1 shot per class to 20), a range of available auxiliary data, and multiple image embedding backbones.
There are several key findings.
First, when auxiliary data is closely related to the target task, \sys is most beneficial in the few-shot setting.
We find here that \sys improves accuracy over BigTransfer~\cite{kolesnikov:bigtransfer} by an average 11.9 percentage points in the one-shot case and 2.13 in the five-shot case.
It also improves over the best-performing semi-supervised learning method by an average of 11.54 and 1.96 points in the one-shot and five-shot cases, respectively.
In an alternative scenario in which only more distantly related auxiliary data is available, \sys significantly outperforms traditional transfer learning and state-of-the-art semi-supervised learning methods in the one-shot and five-shot cases.
In addition, we conduct many ablation experiments to analyze how the multiple components of \sys contribute to its performance.
These results demonstrate that automatically selecting auxiliary data, and then exploiting it in combination with unlabeled data and limited labeled data, can lead to large improvements in accuracy across a range of scenarios.

\section{Related Work}

\sys builds on work in many related areas, including learning with limited labeled data, learning with auxiliary data, and systems for machine learning.

\textbf{Learning with Limited Labeled Data.}
There are many subareas of machine learning focused on learning with limited labeled data, including transfer learning~\citep{pan:tkde10,yosinski:neurips14,weiss:bigdata16,devlin:naacl19} and semi-supervised learning~\citep{chapelle:book06,sohn:fixmatch,chen:simclrbig,pham:metapseudo}.
\sys includes modules based on specific types of transfer learning.
Multi-task learning~\citep{caruana:ml97} trains a single model to solve multiple tasks simultaneously, with the hope that tasks will benefit from information learned from each other.
Zero-shot learning~\citep{xian:pami18} trains models to map semantic descriptions of classes to their examples, so that predictions can be made on novel classes without any previously seen examples.
Few-shot learning~\citep{finn:icml17, snell:neurips17, ravi:iclr17, kolesnikov:bigtransfer} emphasizes transfer learning when only a few labeled examples (often one to five per class) are available in the target domain.
Early few-shot approaches often use meta-learning, which trains the model by simulating many episodes of few-shot learning.
More recently, it has been shown that simply learning a good representation from as much data as possible is often a better alternative ~\citep{tian:eccv20, chen:iccv21}, and pre-trained models like BigTransfer~\citep{kolesnikov:bigtransfer} do well in both few-shot and more conventional transfer learning scenarios.

Other aspects of \sys are connected with specific training strategies.
Our use of unsupervised ensembling of different models trained on different parts of the available data can be viewed as a type of bagging~\citep{breiman:ml96}.
The key difference is that in \sys each ensemble member uses a different learning strategy and types of data (auxiliary, labeled, and unlabeled).
These ensemble members can also be viewed as labeling functions in a weak supervision framework such as Snorkel~\citep{ratner:vldbj20}, but differ in that that we emphasize automatically creating them from auxiliary data, as opposed to hand engineering them.
Knowledge distillation~\citep{hinton:neuripsws14,gou:ijcv21} uses models to train each other, and pseudo labeling can be viewed as a form of knowledge distillation.
Another related area is domain adaptation ~\cite{ganin:jmlr16, tzeng:cvpr17}, which use unlabeled target data to handle the visual domain shift.
There are several few-shot domain adaptation methods~\cite{motiian:neurips17, xu:cvpr19} designed for when few target labels are available. 
However, unlike learning with auxiliary data, in domain adaptation an important assumption is that learning algorithms have access to fully labeled source data with the same label space as the target data.

Other works have considered combining different aspects of these paradigms.
\citet{ren:iclr18,zhou:tr18,sanodiya:ieeeaccess19,wei:cvpr19,yu:cvpr20} and \citet{abuduweili:cvpr21} all consider combining various forms of transfer learning with semi-supervised learning.
\sys differs in that it automatically selects subsets of available auxiliary data for learning.
Recently, \citet{wortsman:arxiv21} showed that a combination of transfer learning and zero-shot learning can improve over either alone.

\textbf{Learning with Auxiliary Data.}
A key aspect of \sys is automatically selecting subsets of auxiliary data to improve accuracy.
Learning with auxiliary data has also been considered in prior work.
\citet{ge:cvpr17} use image similarity to select individual examples for transfer learning.
\citet{cui:cvpr18} pool image similarity scores to select domains of data for transfer learning.
\citet{zhang:eccv18} uses meta-learning to learn to select auxiliary data.
\citet{yan:cvpr20} selects auxiliary data with a mixture of experts trained on the source data.
In contrast, \sys selects auxiliary data by organizing it into a knowledge graph and then operating on the graph.
This approach avoids the need for training any classifiers or computing per-example similarity.

In addition, work on \emph{data discovery}~\citep{deng:cidr17, fernandez:icde18} is complementary to our approach.
Data discovery methods search for and organize potentially useful datasets from data lakes.
We assume that available auxiliary data is organized in a knowledge graph, and data discovery is one way to organize data in this format.

\begin{figure*}[t]
    \centering
    \includegraphics[width=0.9\textwidth]{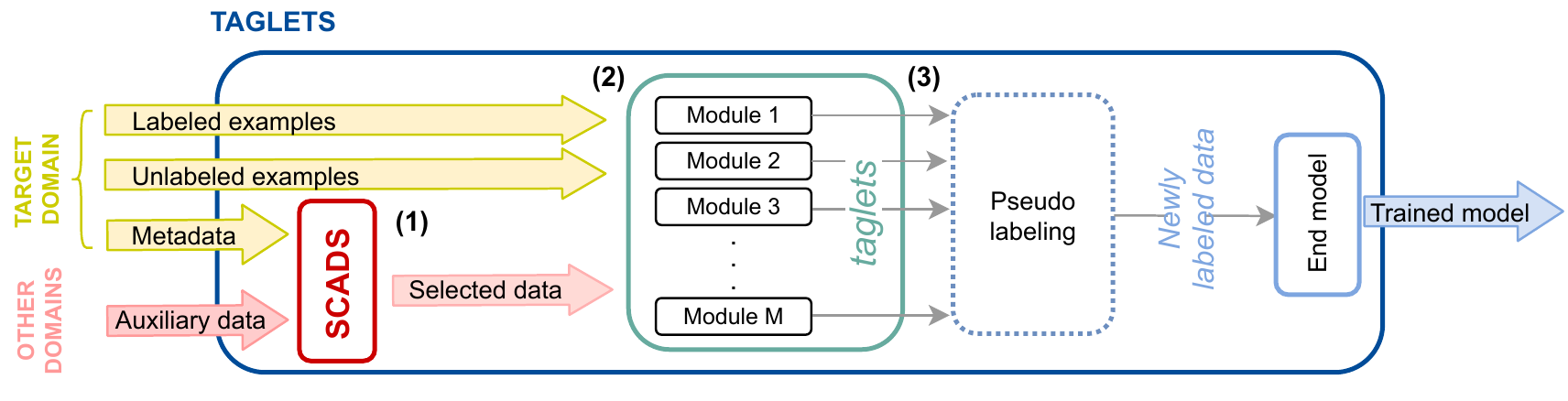}
    \vspace{-1em}
    \caption{\sys's architecture. \textbf{(1)} \scads organizes and finds auxiliary data related to the target classes. \textbf{(2)} Multiple modules are trained to perform pseudo-labeling. \textbf{(3)} In the distillation stage, we pseudo-label the unlabeled data and train the \emph{end model} on both the pseudo-labeled and labeled data.}
    \label{fig:taglets}
\end{figure*}

\textbf{AutoML.}
\sys emphasizes automatic production of classifiers.
Many other works in the field of AutoML have also studied how to automatically produce classifiers and other models, with different focuses.
Many systems focus on pipeline optimization~\citep{thornton:kdd13,sparks:arxiv15,feurer:neurips15,li:jmlr17,kotthoff:jmlr17,shang:sigmod19,smith:sigmod20}.
They generally select algorithms and tune hyperparameters, using either Bayesian optimization~\citep{snoek:neurips12} or multi-armed bandits~\citep{slivkins:arxiv19}, which requires labeled validation data.
Other systems automate the feature engineering process~\citep{kanter:dsaa15,khurana:icdmw16}.
These approaches are all potentially complementary with \sys, but they require labeled validation data.
In contrast, \sys is targeted at few-shot scenarios.
We find in our experiments that a single set of modules and hyperparameters works well across tasks, and our focus is instead on automating the process of exploiting the full spectrum of auxiliary, unlabeled, and limited labeled data.

\section{\sys Architecture}\label{sec:architecture}

\sys learns a task $T$ and returns a servable classifier $\model: \mathcal{X}_T \to \mathcal{Y}_T$ by exploiting ecosystems of datasets and models simultaneously. 

The system has access to a spectrum of data:
\begin{enumerate}
    \item Limited labeled examples $\labset$,
    \item Task unlabeled data $\unlabset$,
    \item Auxiliary data $\auxset$, from various domains $S$,
\end{enumerate}

and it consists of the following components (\Cref{fig:taglets}):
\begin{enumerate}
    \item \textbf{A structured collection of annotated datasets (\scads).} 
    We organize auxiliary data in a repository according to a knowledge graph called a \scads~(\Cref{sec:scads}).
    We use relations among classes to automatically \emph{select} subsets of task-related auxiliary data $\selectedauxsetname$.

    \item \textbf{Training Modules.} 
    The second component of the system incorporates independent transfer, semi-supervised, and zero-shot learning modules, specifically tailored to exploit task-related auxiliary data~(\Cref{sec:modules}).
    The output of each module is a classifier, which we call a \emph{taglet}. 
    A taglet takes an input example $x \in \xtarget$ and returns a label $y \in \ytarget$. 
    We denote the collection of taglets $\tagletcollection$.

    \item \textbf{Distillation stage.}
    The predictions of different taglets are combined to create \emph{pseudo labels} for the unlabeled examples, which are used as targets to learn the end model ~\Cref{sec:endmodel}.
\end{enumerate}

\begin{figure*}[t]
    \centering
    \includegraphics[width=1\textwidth]{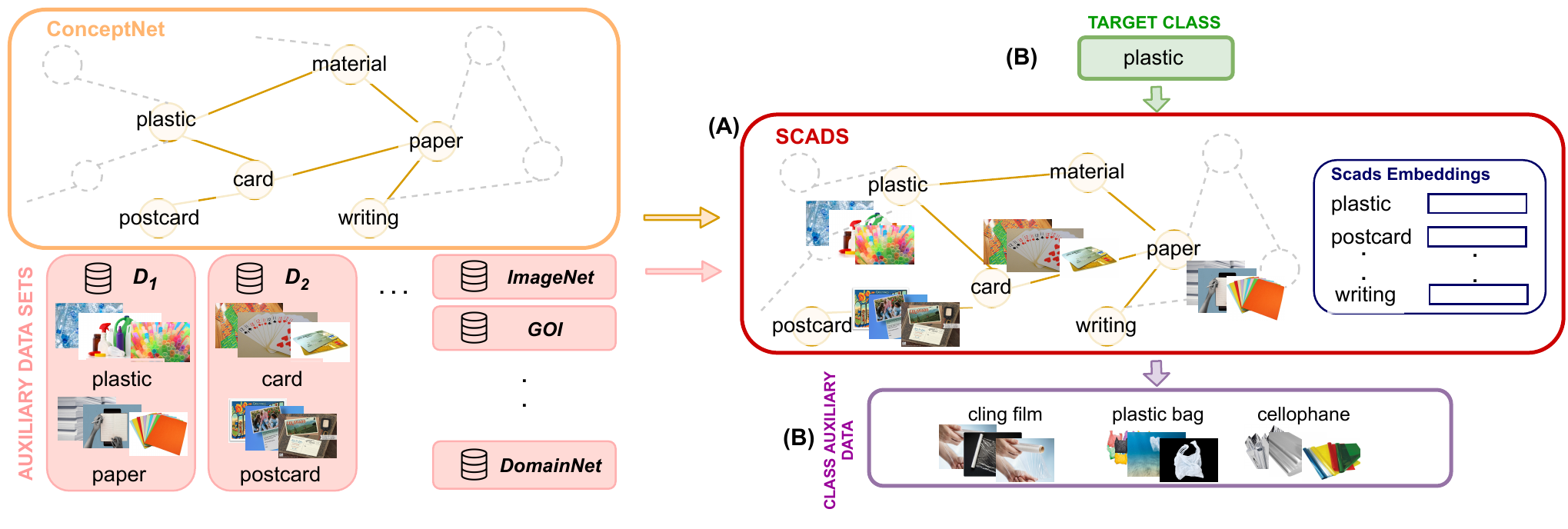}
    \vspace{-2em}
    \caption{\scads is built by aligning auxiliary categories with concepts (nodes) in a knowledge graph such as ConceptNet. \textbf{(A)} All images of a category belong to the same corresponding concept, and each concept is associated with an embedding, i.e., \emph{scads embedding}, which can be used to determine the relations among concepts. \textbf{(B)} Given a query class \texttt{plastic}, \scads retrieves the most similar concepts, according to graph-based similarity.}
    \label{fig:scads}
\end{figure*}
\subsection{SCADS}\label{sec:scads}

The goal of \scads is to help identify a relevant subset of auxiliary data for a target task.
In general, categories described in annotated datasets can  overlap or relate to each other. 
For instance, given two datasets, both gathering images of \emph{postcards}, we need to query them separately to extract all the corresponding examples.
Further, when two datasets collect images of related categories, e.g., one contains images of \emph{paper}, and the other of \emph{cards} and \emph{postcards}, we need to automatically identify that these examples are related to associate them.
We therefore design \scads to efficiently retrieve examples from related categories using semantic relations among them.
\Cref{fig:scads} shows an example of a subset of a \scads.

\textbf{Construct \scads.}
Users can construct a \scads for their domain by joining all of their data to a \emph{common sense knowledge graph}.
Common sense knowledge graphs are powerful resources that make explicit many facts about the world by stating relationships among millions of concepts.
We propose using them as the underlying structure of \scads. 
In particular, we choose ConceptNet~\cite{speer:conceptnet}, a common sense knowledge graph $\graph=(\nodes, \edges)$ whose nodes $\nodes$ are words and phrases of natural language (i.e., concepts), and edges $\edges$ express their relations.
To build \scads, we map the classes $\class \in \yaux$ in the auxiliary data and concepts $\concept \in Q$, such that all the images of the same class are associated to the corresponding node in $\graph$ (\Cref{fig:scads} A).
We denote $Q_{\yaux}$ the subset of $Q$ matching with $\yaux$.
\scads enables end-users to easily install and remove any available annotated dataset.
Its extensibility makes possible to unlock the potential of datasets that otherwise could not as easily be used as standalone sources of information.
This joining process requires a one-time labor cost, but we find that it is small and can be automated for large datasets like ImageNet-21k using the many links in ConceptNet to external resources like WordNet and Wikipedia.

\textbf{Select auxiliary data}. 
Joining auxiliary data and ConceptNet defines a rich set of semantic relations on top of examples, and we want to exploit them to select a relevant subset of auxiliary data for the target task.
Typical approaches to find data related to the target classes rely on visual similarity~\cite{cui:cvpr18,yan:cvpr20}.
However, such methods can be computationally expensive, by involving pairwise calculations.
In our setting, we look for task-related data among enormous amounts of auxiliary data (e.g., ImageNet-21k), so visual pairwise-comparisons become intractable. 
Thus, we leverage \scads underlying structure and use graph-based semantic similarity between concepts in $Q_{\yaux}$ and a target class $\class$, to find the subset of concepts mostly related to $\class$. Further, from this subset we query \scads to get all the corresponding images.
With $|Q_{\yaux}| << |\auxsetname|$, our approach is efficient and scales well with increasing amount of auxiliary data in \scads.
In addition, we note that our approach is robust over visual domain shifts, because it does not involve any operation on the images.
Computing graph-based similarities requires the definition of \emph{SCADS embeddings}, i.e., a representation of the nodes in \scads.
Since the underlying graph of \scads is ConceptNet, we use its pre-computed embeddings~\cite{speer:conceptnet}, which express both the network topology and word embeddings learned from text.
In~\Cref{app:scads}, we provide details about the technique to compute these embeddings.
\begin{exmp}\label{example:scads}
\textit{Suppose our task is to classify surfaces according to their materials, and consider the target class \texttt{plastic} (\Cref{fig:scads} B).
From the \emph{scads embeddings} we get the vector $\scadsemb{plastic}$ and use the cosine similarity to find the top-$\numrelated$ closest concepts in $\nodes$, i.e., \texttt{cling film, plastic bag, cellophane}.}
\end{exmp}
For each of the $\numrelated$ similar concepts, \scads retrieves from the corresponding nodes a set of $\numshots$ images.
In this way, having $|\ytarget|=C$ target classes, we get the set $\selectedauxsetname$ consisting of $C \cdot (\numrelated \cdot \numshots)$ selected auxiliary examples.
\scads provides flexibility for compute budgets by allowing users to fix the size of the selected auxiliary data $\selectedauxsetname$ by setting threshold parameters for the number of task-related concepts $\numrelated$ and the number of associated examples $\numshots$.
These thresholds allow the training time of modules (\Cref{sec:modules}) to remain constant even when the total size of auxiliary data $\auxsetname$ is large.

To align target classes with a \scads, we can use a heuristic match with existing nodes or manual selection.
Even in circumstances where a given target class does not align well with any node in ConceptNet, users can manually add nodes and edges to existing concepts.

\begin{exmp}\label{example:not_in_scads_body}
\textit{While developing a model to recognize objects in a grocery store (\Cref{sec:datasets}), we found target classes representing specific products without a counterpart in \scads, e.g. \texttt{oatghurt}. 
We handle this scenario by creating a new node \texttt{oatghurt} and linking it to existing characterizing concepts, e.g., \emph{yoghurt, carton}, and \emph{oat milk}.}
\end{exmp}
In~\Cref{example:not_in_scads_body}, the target class \texttt{oatghurt} is not included in the \emph{scads embedding}.
Thus, we compute an \emph{approximation embedding} $\scadsemb{q} \approx \sum_{j \in P} w_j \cdot e_j$, where $P$ is the set of approximation terms that share a prefix as long as possible with the given term, and $w_i$ is a term's weight that we set to $\frac{1}{|P|}$.
We note that the possibility of modifying the graph underlying \scads emphasizes the \emph{extensibility} of the component.

\subsection{Modules}\label{sec:modules}

In \sys, we include four methods in the areas of transfer, semi-supervised and zero-shot learning, and adapt them to fully exploit the knowledge collected in \scads.
We call a method tailored to use \scads a \emph{module} and its output (i.e., a trained model) a \emph{taglet}. 
Formally, a module $\module$ takes in input data among $\labsetname, \unlabsetname$, and $\selectedauxsetname$, and returns a taglet $\taglet{m} : x \mapsto y \in [0,1]^{|\ytarget|}$, such that $\sum_{c=1}^{|\ytarget|} y_c = 1$. 
Modules are independently trained, and their predictions are ensembled to take advantage of their diversity (\Cref{sec:endmodel,sec:intermediate_vs_final}).
This modular framework is extensible, as other methods can be incorporated on top of the ones we develop here.

\subsubsection{Transfer Module}\label{sec:transfer}

In the scenario of learning with limited labeled data, fine-tuning a pre-trained model $\encoder$ on the target task is the common way of transferring knowledge across domains~\cite{donahue:decaf,yao:transfer, oquab:transfer, razavian:offshelf}. 
The \emph{transfer module} fine-tunes sequentially on task-related auxiliary and labeled data to leverage the knowledge in \scads. 
We train the \emph{transfer module} with an off-the-shelf model $\encoder$ as follows:
\begin{enumerate}
    \item We extract the set $\selectedauxsetname$ of selected auxiliary data from \scads with the strategy presented in~\Cref{example:scads}, and note that $|\selectedauxsetname| = C \cdot (\numrelated \cdot \numshots)$.
    \item \emph{Intermediate phase:} We learn the new weights $\weightsencoder'$ of $\encoder$  fine-tuning its backbone on $\selectedauxsetname$, by minimizing the cross entropy loss:
    \begin{equation}\label{eq:finetuning}
        \loss{inter} = \frac{1}{|\selectedauxsetname|}\sum_{i=1}^{|\selectedauxsetname|}  \sum_{c=1}^{\numrelated C} - y_i^{(c)} \cdot \log\encoder_{\weightsencoder'}(x_i)^{(c)}.
    \end{equation} 
    \item We train the final model by fine-tuning $\encoder_{\weightsencoder'}$ on the labeled examples $\labsetname$, and get the new weights $\weightsencoder''$. 
    \begin{equation}
         \loss{target} = \frac{1}{|\labsetname|}\sum_{i=1}^{|\labsetname|}  \sum_{c=1}^{C} - y_i^{(c)} \cdot \log\encoder_{\weightsencoder''}(x_i)^{(c)},
    \end{equation} 
    where $y_i^{(c)}$ is 1 if $x \in c$, otherwise 0.
\end{enumerate}

\subsubsection{Multi-task Module}\label{sec:multitask}

An alternative way to transfer knowledge across domains is through multi-task learning~\citep{caruana:ml97} that consists of learning several related tasks simultaneously.
The task relatedness has been vaguely defined, and we propose to specify it as the semantic similarity between the classes of the target and auxiliary classification tasks.
The \textit{multi-task module} forms an auxiliary classification task out of selected auxiliary data, most semantically similar to the target data, and jointly learns the target task on $\labsetname$ and the auxiliary task on $\selectedauxsetname$. 
In particular, we extract the set of selected auxiliary data $\selectedauxsetname$, select a pre-trained model $\encoder$, and learn its weights $\weightsencoder'$, by optimizing over a weighted sum of cross entropy losses of the target and and auxiliary tasks: 
\begin{equation}
     \loss{joint} = \loss{target} + \lambda \loss{aux},
\end{equation} 
where $\lambda \in \mathbb{R}$ controls the influence of auxiliary task in the joint training, and the losses are:
\begin{equation}
     \loss{target} = \frac{1}{|\labsetname|}\sum_{i=1}^{|\labsetname|}  \sum_{c=1}^{C} - y_i^{(c)} \cdot \log\encoder_{\weightsencoder'}(x_i)^{(c)}.
\end{equation} 
\begin{equation}
    \loss{aux} = \frac{1}{|\selectedauxsetname|}\sum_{i=1}^{|\selectedauxsetname|}  \sum_{c=1}^{\numrelated C} - y_i^{(c)} \cdot \log\encoder_{\weightsencoder'}(x_i)^{(c)}.
\end{equation}

\subsubsection{FixMatch Module}\label{sec:fixmatch}
Since \sys has access to unlabeled data, we include FixMatch~\cite{sohn:fixmatch}, an inductive semi-supervised learning algorithm that learns a model $\encoder$ by employing pseudo labeling and consistency regularization. 
The pseudo labeling procedure uses the model $\encoder$ to label each $x \in \mathcal{U}$ and then trains with the newly labeled instances as targets, assuming the label confidence values are larger than a threshold $\tau$. 
Additionally, consistency regularization uses unlabeled examples to encourage $\encoder$ to be resistant to information-preserving transformations of input images.

Under the regime of limited labeled data, naively applying consistency-based semi-supervised learning algorithms can result in poor performance due to confirmation bias \cite{wang2021selftuning}. 
We address this by initializing $\encoder$ with a pre-trained model that is further fine-tuned on the auxiliary data $\selectedauxsetname$ extracted from \scads.
The \emph{FixMatch module} minimizes the objective:
\begin{equation*}
    \loss{fix} = \sum_{\substack{u \in \mathcal{U} \\ u_a, u_b = \alpha(u)}} \mathbbm{1}(\max(\encoder_{\weightsencoder'}(u_a)) \geq \tau) H(u_b, \encoder_{\weightsencoder'}(u_a)),
\end{equation*}
where $H$ is the cross entropy function and $\alpha$ is stochastic function, such as a random rotation, that returns two augmented versions of a single input, i.e., $u_a$ and $u_b$.

\subsubsection{ZSL-KG Module}\label{sec:zslkg}

Finding generalized representations of target classes with minimal data can be hard, especially with only one labeled example per class.
To complement the first three modules, we include a zero-shot learning module that makes predictions without any labeled data for the target task.
Instead, it uses only the knowledge graph in \scads.
Among the available methods in this subfield, we choose ZSL-KG~\citep{nayak:arxiv20}, an inductive model for generating class representations from common sense knowledge graphs. 
The \emph{ZSL-KG module} uses an instance of ZSL-KG $Z$ pre-trained on ConceptNet and ImageNet 1k to learn the representation of target classes.
Given a node in a knowledge graph mapped to the target class, ZSL-KG predicts the class' weights vector that can be used in a classification head of an off-the-shelf encoder $\encoder$.

In particular, the module works as follows:
\begin{enumerate}
    \item We take the graph $\graph$ underlying \scads, and the concept $\concept$ associated to the target class $\class$, and pass them to the pre-trained graph neural network $Z$, to generate the class representation $z_{\class} = Z(\concept, \graph)$.
    \item We generate the class representation $z_{\class}$ for all the target classes and plug them into $\encoder$ as the weights of the classification head.
\end{enumerate}

\subsection{Distillation stage}\label{sec:endmodel}
We distill the knowledge of individual modules into one servable \emph{end model}.
We \emph{ensemble} the predictions of all $\taglet{m} \in \tagletcollection$ on the unlabeled data to produce \emph{pseudo labels} capturing the knowledge of all taglets.

\textbf{Ensembling.} Given an example $x$:  
\begin{enumerate}
    \item Taglets in $\tagletcollection$ return probability vectors $\hat{y}^{t_m} \in [0,1]^{C}$.
    \item We concatenate vectors $\hat{y}^{t_1}, \hat{y}^{t_2}, \cdots, \hat{y}^{t_{|\tagletcollection|}}$ into a vote matrix $\vote \in [0,1]^{|\tagletcollection| \times C}$.
    \item We aggregate the prediction vectors in $\vote$ and compute the \emph{soft}-pseudo-label:
        \begin{equation}\label{eq:aggregation}
            p_x = \frac{1}{|\tagletcollection|}\sum_{t=1}^{|\tagletcollection|}\vote_{t},
        \end{equation}
    where the scaling factor $|\tagletcollection|$ ensures $p_x \in [0,1]^{C}$ is a probability vector.
\end{enumerate}

The benefits of the ensemble as a means of exploiting the \emph{taglets}' diversity are discussed in~\Cref{sec:intermediate_vs_final}.

\textbf{End model.} 
After we assign to all $x \in \unlabsetname$ pseudo labels, we use them, along with labeled data $\labsetname$, to train the end model $\model$.
We train $\model$ fine-tuning a pre-trained model on the pseudo-labeled data $\mathcal{P}$ and $\labsetname$, by optimizing the soft cross entropy loss:
\begin{equation}
 \loss{final} = \frac{1}{|\mathcal{P} \cup \labsetname|}\sum_{j=1}^{|\mathcal{P} \cup \labsetname|}  \sum_{c=1}^{C} - p_j^{(c)} \cdot \log\model_{\theta'}(x_j)^{(c)}.
\end{equation} 
We emphasize the importance of producing a single, servable model when service-level agreements are required.

\section{Experimental evaluation}\label{sec:experiments}

In this section, we evaluate \sys on four different image classification tasks with the goal of showing
(1) the predictive accuracy of \sys under different circumstances,  
(2) the benefits of \scads, and (3) the benefits of combining multiple learning techniques. 
The code to reproduce the results presented in the reminder of the section is openly available at \url{https://github.com/BatsResearch/piriyakulkij-mlsys22-code}.

\subsection{Datasets}\label{sec:datasets}

We refrain from using standard semi-supervised learning datasets, such as ImageNet and CIFAR derivatives, which are general, natural image classification datasets. 
Instead, we turn to datasets designed for specific applications, not necessarily describing general visual domains.
This choice is driven by two main reasons: (1) massive general, natural image datasets, including Google Open Images~\cite{kuznetsova:openimagev4} and ImageNet-21k~\cite{deng:imagenet22k} are publicly available, so practitioners can rely on them to get enough data to solve target tasks in a general, natural domain,
and (2) \sys is designed to solve tasks for specific applications. 

\textbf{Flickr Material Database}~\cite{sharan:fmd} is a benchmark for material recognition and captures the natural range of material appearances. It has 1,000 color photographs of surfaces uniformly belonging to 10 common material categories like \emph{fabric, foliage}, and \emph{glass}. 
To reduce the chances that simple, low-level information (e.g., color) can be used to distinguish material categories, the dataset intentionally includes diversity in the same class.
A practical application is to support waste sorting and recycling.

\textbf{Office-Home dataset}~\cite{venkateswara:officehome} consists of 65 object categories, like \emph{keyboard, desk lamp}, and \emph{fan},  found typically in office and home settings.
The images are from four different domains, among which we consider \emph{clipart} including clipart images, and \emph{product} gathering images without background, and build the \textbf{OfficeHome-Clipart} and the \textbf{OfficeHome-Product} datasets. 
The two contain a minimum of 39 and 38 images per class respectively. 
This dataset requires recognizing daily life objects in non-natural visual domains.
A practical application is understanding images on the Web, such as product catalogs and illustrations.

\textbf{Grocery Store dataset}~\cite{klasson:grocery} consists of natural images of grocery items taken with a smartphone camera in different grocery stores. 
The dataset contains 5,125 natural images from 42 coarse-grained classes of fruits, vegetables, and carton items, with the minimum number of images per class being 18. 
The dataset highlights the extensability of \sys because two target classes, i.e., \texttt{oathgurt} and \texttt{soygurth}, do not appear as concepts in ConceptNet. 
To address this issue, we modify \scads by adding new nodes and manually connect them and existing nodes (see~\Cref{example:not_in_scads_body}).  s
A practical application is assistive technology for people with vision impairments.

\begin{figure*}[t]
    \centering
    \includegraphics[width=0.9\textwidth]{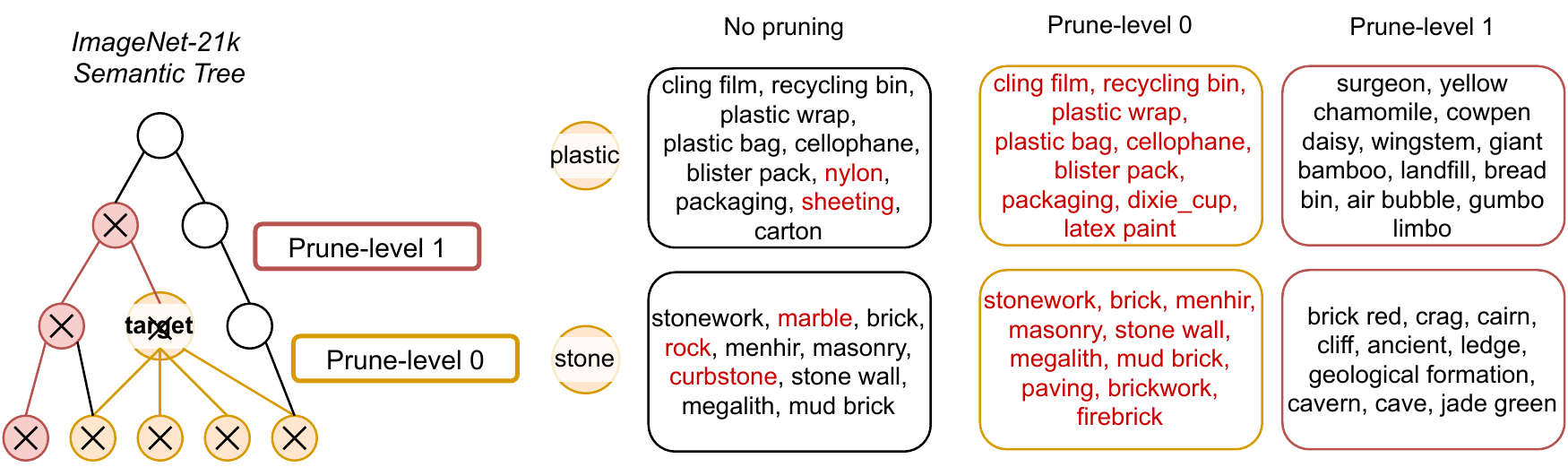}
    \caption{Two examples that demonstrate how pruning affects the retrieval of concepts related to a target class. Given the target class \texttt{plastic}, we find the list of ten most related concepts in \scads, without pruning. We highlight in red the concepts that are pruned from \scads when we remove \texttt{plastic} and its children from the ImageNet-21k semantic tree (prune level 0). We do the same for the next level of pruning. At higher level of pruning, the remaining concepts are more general or more distantly related to the target task.}
    \label{fig:pruning}
\end{figure*}

\subsection{Baselines}\label{sec:baselines}

We compare \sys to state-of-the-art transfer and semi-supervised learning methods.

\textbf{Fine-tuning}.
Our system is designed to transfer knowledge across its components, starting from auxiliary data related to the target classes.
We compare \sys with \textbf{BigTransfer} (\textbf{BiT})~\cite{kolesnikov:bigtransfer} a model pre-trained on ImageNet-21k~\cite{deng:imagenet22k} that uses a ResNet-50-v2 as backbone~\cite{he:identity}, and stands out for its performances when fine-tuned with few labeled examples per class.
To study situations in which a model pre-trained on all available auxiliary data is not available, we also compare with \textbf{ResNet-50}~\cite{kaiming:cvpr16} pre-trained on ImageNet-1k~\cite{russakovsky:imagenet}, a standard image encoder.

\textbf{Distilled Fine-tuning}.
Since \sys has a distillation stage that uses unlabeled data, we also run distilled versions of fine-tuning.
In particular, we first fine-tune the backbone of either BiT (ImageNet-21k) or ResNet-50 (ImageNet-1k) on the available labeled examples, then produce pseudo labels for $\unlabsetname$, and eventually fine-tune either BiT or ResNet-50 using the newly labeled examples as targets.

\textbf{Semi-supervised Methods}. 
We compare \sys against three state-of-the-art semi-supervised learning algorithms: FixMatch \cite{sohn:fixmatch}, Meta Pseudo Labels \cite{pham:metapseudo}, and SimCLRv2 \cite{chen:simclrbig}. 
While these methods typically use randomly initialized ResNet encoders, we use pre-trained encoders in our experiments as they give better performance with the limited amount of both labeled and unlabeled data.
\textbf{FixMatch} combines consistency regularization and pseudo labeling, as explained in~\Cref{sec:fixmatch}. 
For fair comparisons, we implement FixMatch using either ResNet-50 (ImageNet-1k) or BiT (ImageNet-21k) as pretrained encoders. \textbf{Meta Pseudo Labels} trains a student network on unlabeled data using pseudo labels generated from a teacher network.
Unlike in traditional approaches to pseudo labeling, feedback from the student's performance is also used to adapt the teacher. 
After teacher-student training, the student is fine-tuned on labeled data to reduce the impact of confirmation bias.
Similar to FixMatch, the teacher network uses either ResNet-50 (ImageNet-1k) or BiT (ImageNet-21k) as a pretrained encoder.
The student network, however, only uses ResNet-50 (ImageNet-1k) as a pretrained encoder to prevent overfitting. \textbf{SimCLRv2} uses data augmentation techniques combined with a contrastive loss function to pre-train an encoder with little to no supervision. 
This approach has exhibited strong performance when trained on large unlabeled datasets, such as ImageNet-1k. 
However, in our experiments, we found that the performance of SimCLRv2 deteriorates significantly when trained on smaller datasets.
Consequently, we do not include this method in our results.

\subsection{System and Experiments Set Up}\label{sec:sys_spec}

\begin{table*}[h]
    \centering
    \resizebox{\linewidth}{!}{
    \begin{tabular}{lccccccc}
    \toprule
    Method & Backbone & \multicolumn{3}{c}{OfficeHome-Product} & \multicolumn{3}{c}{OfficeHome-Clipart}\\
    \midrule
    & & 1-shot & 5-shot & 20-shot & 1-shot & 5-shot & 20-shot\\
    \cmidrule(lr){3-5} \cmidrule(lr){6-8}
    Fine-tuning & BiT (ImageNet-21k) & 57.28 $\pm$ 5.20 & 84.41 $\pm$ 1.17 & \textbf{90.67 $\pm$ 1.10} & 30.36 $\pm$ 2.54 & 63.44 $\pm$ 2.10 & \textbf{77.13 $\pm$ 1.34} \\
    Fine-tuning (Distilled) & BiT (ImageNet-21k) & 58.21 $\pm$ 4.56 & 85.03 $\pm$ 0.22 & \textbf{92.00 $\pm$ 0.38} & 31.74 $\pm$ 1.96 & 62.92 $\pm$ 1.99 & \textbf{78.26 $\pm$ 0.88}\\
    FixMatch & BiT (ImageNet-21k) & 49.03 $\pm$ 2.54 & 83.03 $\pm$ 2.97 & \textbf{92.21 $\pm$ 0.44} & 30.67 $\pm$ 5.44 & 64.77 $\pm$ 1.75 & \textbf{79.95 $\pm$ 2.30} \\
    Meta Pseudo Label & BiT (ImageNet-21k) & 59.95 $\pm$ 0.44 & 81.53 $\pm$ 1.01 & \textbf{89.90 $\pm$ 0.88} & 31.74 $\pm$ 2.1 & 62.92 $\pm$ 2.13 & 76.05 $\pm$ 1.72 \\
    TAGLETS & BiT (ImageNet-21k) & \textbf{70.92 $\pm$ 1.67} & \textbf{86.31 $\pm$ 1.01} & \textbf{92.36 $\pm$ 3.18} & \textbf{46.97 $\pm$ 2.46} & \textbf{67.64 $\pm$ 3.34} & \textbf{78.62 $\pm$ 2.76} \\
    \midrule
    Fine-tuning & ResNet-50 (ImageNet-1k) & 32.51 $\pm$ 3.83 & 63.44 $\pm$ 7.45 & 81.13 $\pm$ 0.22 & 16.26 $\pm$ 2.87 & 43.49 $\pm$ 0.58 & 68.56 $\pm$ 2.24 \\
    Fine-tuning (Distilled) & ResNet-50 (ImageNet-1k) & 32.72 $\pm$ 3.67 & 66.87 $\pm$ 5.60 & 83.85 $\pm$ 2.68 & 15.85 $\pm$ 0.66 & 44.26 $\pm$ 2.34 & \underline{72.00 $\pm$ 2.32}\\
    FixMatch & ResNet-50 (ImageNet-1k) & 39.28 $\pm$ 3.34 & 72.87 $\pm$ 1.45 & 85.03 $\pm$ 1.81 & 20.51 $\pm$ 0.96 & 49.69 $\pm$ 3.31 & 66.56 $\pm$ 1.23 \\
    Meta Pseudo Label & ResNet-50 (ImageNet-1k) & 47.38 $\pm$ 2.98 & 76.67 $\pm$ 1.75 & \underline{87.38 $\pm$ 1.75} & 22.77 $\pm$ 3.99 & 57.25 $\pm$ 4.64 & \underline{75.17 $\pm$ 1.22} \\
    TAGLETS & ResNet-50 (ImageNet-1k) & \underline{67.64 $\pm$ 3.61} & \underline{83.54 $\pm$ 2.89} & \underline{88.77 $\pm$ 2.89} & \underline{43.44 $\pm$ 4.46} & \underline{63.90 $\pm$ 1.17} & \underline{75.54 $\pm$ 4.34} \\
    \midrule
    TAGLETS prune-level 0 & ResNet-50 (ImageNet-1k) & \underline{65.33 $\pm$ 1.23} & \underline{81.23 $\pm$ 0.00} & \underline{88.62 $\pm$ 1.38} & \underline{41.49 $\pm$ 1.54} & 62.21 $\pm$ 1.89 & \underline{75.79 $\pm$ 3.11}  \\
    TAGLETS prune-level 1 & ResNet-50 (ImageNet-1k) & 57.28 $\pm$ 1.89 & 76.21 $\pm$ 1.54 & \underline{86.67 $\pm$ 0.58} & 36.87 $\pm$ 2.17 & 55.33 $\pm$ 4.19 & \underline{72.72 $\pm$ 0.96}\\
    \bottomrule
    \end{tabular}}
    \caption{Accuracy (\%) of \sys and baselines on \emph{OfficeHome Product} and \emph{Clipart} (\texttt{Split 0}) for increasing number of labeled examples (1, 5, and 20 shots), different backbones (ResNet-50 trained on ImageNet-1k and BiT trained on ImageNet-21k) and levels of pruning (no-pruning, 0 and 1). We bold and underline the best performances, and those within their 95\% confidence intervals, carried out using \textbf{BiT} and \underline{ResNet} backbones, respectively. Measurements are repeated for three seeds and we report the 95\% confidence intervals.}
\label{tab:acc:officehome}
\end{table*}

\emph{System}. 
On top of ConceptNet, we use ImageNet-21k as the auxiliary dataset in \scads because it offers a comprehensive overview of categories covering common knowledge helpful to solve our tasks. We point out that our choice can be combined with other annotated datasets potentially useful for the target task.
Throughout the experiments, we vary the  backbones of the modules (i.e., pre-trained models $\encoder$) choosing either ResNet-50  trained on ImageNet-1k or BiT trained on ImageNet-21k. 
The two options represent how the backbone can be either pre-trained on parts or all of the auxiliary data, though \sys's users can also plug in any backbone of their choices that is not necessarily pre-trained on data installed on \scads. 
For each backbone, the hyperparameters of \sys are fixed throughout our experiments (see~\Cref{app:training} for details). 
The good quality of predictions obtained across datasets without hyperparameter tuning highlights the robustness of the system.

\emph{Experiments}. 
For each of the datasets, we randomly generate a train and a test set, if the latter is not predetermined (\Cref{app:experiments}).
In the train set, randomly pick and label 1, 5, or 20 examples (i.e., shots) per class, and use the rest as unlabeled examples. 
We run the system on three different train-test splits on each dataset (\Cref{app:experiments}). Due to space constraints, we report results of one split and gather the others in~\Cref{app:additional_results}.
The trends are consistent across all three splits.

\emph{Pruning}. We  \emph{prune} \scads to simulate the scenario in which only more distantly related auxiliary data is available, and investigate if the relatedness between auxiliary data and target classes is a crucial factor to boost \sys performance.
Pruning \scads consists in progressively removing concepts closely related to the target classes. 
In particular, we take the semantic tree $H$ defined over the categories in ImageNet-21k, i.e., the WordNet hierarchy~\cite{fellbaum:wordnet}, and for the target class $\class$ remove: $\class$ and all its descendants to obtain \textbf{prune-level 0}. Further, we remove $\class$'s parent and the parent's subtree for \textbf{prune-level 1}.
See~\Cref{fig:pruning}.

\begin{table*}[ht]
    \centering
    \resizebox{\linewidth}{!}{
    \begin{tabular}{lcccccc}
    \toprule
    Method & Backbone & \multicolumn{2}{c}{Grocery Store Dataset} & \multicolumn{3}{c}{Flickr Material Dataset}\\
    \midrule
    & & 1-shot & 5-shot & 1-shot & 5-shot & 20-shot\\
    \cmidrule(lr){3-4} \cmidrule(lr){5-7}
    Fine-tuning & BiT (ImageNet-21k) & 51.32 $\pm$ 3.23 & 85.16 $\pm$ 2.41 & 52.67 $\pm$ 2.35 & 70.07 $\pm$ 1.52 & \textbf{85.40 $\pm$ 0.99}\\
    Fine-tuning (Distilled) & BiT (ImageNet-21k) & 52.42 $\pm$ 1.03 & 86.06 $\pm$ 1.81 & 57.53 $\pm$ 10.95 & 73.27 $\pm$ 1.03 & \textbf{86.80 $\pm$ 1.31}\\
    FixMatch & BiT (ImageNet-21k) & 48.58 $\pm$ 13.41 & 87.61 $\pm$ 4.82 & 19.07 $\pm$ 31.91 & 60.67 $\pm$ 5.03 & \textbf{84.40 $\pm$ 2.28} \\
    Meta Pseudo Label & BiT (ImageNet-21k) & 54.51 $\pm$ 6.22 & 84.02 $\pm$ 1.30 & 52.73 $\pm$ 5.82 & 69.80 $\pm$ 2.24 & 79.33 $\pm$ 2.24\\
    TAGLETS & BiT (ImageNet-21k) & \textbf{61.60 $\pm$ 2.90} & \textbf{88.91 $\pm$ 1.07} & \textbf{68.07 $\pm$ 5.76} & \textbf{75.20 $\pm$ 1.72} & \textbf{84.53 $\pm$ 2.29}\\
    \midrule
    Fine-tuning & ResNet-50 (ImageNet-1k) & 30.25 $\pm$ 4.08 & 70.60 $\pm$ 3.25 & 28.20 $\pm$ 3.48 & 47.20 $\pm$ 3.94 & 60.47 $\pm$ 8.52\\
    Fine-tuning (Distilled) & ResNet-50 (ImageNet-1k) & 30.64 $\pm$ 4.33 & 74.63 $\pm$ 2.44 & 29.20 $\pm$ 7.32 & 49.80 $\pm$ 3.44 & 66.27 $\pm$ 4.51\\
    FixMatch & ResNet-50 (ImageNet-1k) & 32.14 $\pm$ 4.72 & 70.77 $\pm$ 0.75 & 37.73 $\pm$ 13.97 & 61.00 $\pm$ 6.72 & 74.13 $\pm$ 1.52  \\
    Meta Pseudo Label & ResNet-50 (ImageNet-1k) & 34.68 $\pm$ 3.11 & 75.16 $\pm$ 1.57 & 37.99 $\pm$ 13.52 & \underline{68.00 $\pm$ 3.47} & \underline{77.13 $\pm$ 1.25}\\
    TAGLETS & ResNet-50 (ImageNet-1k) & \underline{51.76 $\pm$ 3.48} & \underline{84.35 $\pm$ 0.92} &  \underline{55.87 $\pm$ 4.05} & \underline{68.27 $\pm$ 2.45} & 74.53 $\pm$ 2.74 \\
    \midrule
    TAGLETS prune-level 0 & ResNet-50 (ImageNet-1k) & 46.32 $\pm$ 5.01 & 81.88 $\pm$ 1.27 & \underline{53.40 $\pm$ 4.42} & \underline{67.60 $\pm$ 3.48} & \underline{74.47 $\pm$ 1.88} \\
    TAGLETS prune-level 1 & ResNet-50 (ImageNet-1k) & 37.68 $\pm$ 6.64 & 75.75 $\pm$ 1.84 & \underline{55.00 $\pm$ 8.18} & \underline{66.00 $\pm$ 5.73} & \underline{72.93 $\pm$ 1.60} \\
    \bottomrule
    \end{tabular}}
    \caption{Accuracy (\%) of \sys and baselines on \emph{Grocery Store} and \emph{Flicker Material Datasets} (\texttt{Split 0}) for increasing number of labeled examples (1, 5, and 20 shots), different backbones (ResNet-50 trained on ImageNet-1k and BiT trained on ImageNet-21k) and levels of pruning (no-pruning, 0 and 1). We bold and underline the best performing models, and those within their 95\% confidence intervals, carried out using \textbf{BiT} and \underline{ResNet} backbones, respectively. Measurements are repeated for three seeds, and we report the 95\% confidence intervals.}
\label{tab:acc:gs_and_fms}
\end{table*}

\subsection{Results}\label{sec:results}

We evaluate \sys, addressing the following questions: 
\begin{enumerate}
    \item
    How does the system's accuracy compare with state-of-the-art methods, across varying amounts of labeled examples? (\cref{sec:accuracy}) 
    \item 
    How does auxiliary data improve the modules' performances? (\cref{sec:pruning_accuracy})
    \item 
    How does module ensembling affect the overall system's performance? (\cref{sec:intermediate_vs_final}) 
\end{enumerate}

In~\Cref{sec:intro}, we summarized the accuracy of \sys by averaging the performances over all the tasks and on the three splits. 
In this section, we discuss results on \texttt{split 0} which, according to the 95\% confidence intervals, are consistent with those of \texttt{split 1} and \texttt{split 2} that we report in~\Cref{app:additional_results}.

\subsubsection{Accuracy with Limited Labeled Data}\label{sec:accuracy}

\Cref{tab:acc:officehome} and \Cref{tab:acc:gs_and_fms} show that \sys achieves the best accuracy in 1-shot and 5-shots compared to the existing state-of-the-art methods.
\sys significantly improves over transfer learning methods by 11.9\% and 2.7\%, on average, in the 1 and 5-shots settings, using BiT as the backbone.
Our system is also competitive with the best-performing transfer learning baseline (Distilled Fine-tuning on BiT) in the 20-shots setting.
Similarly, \sys outperforms semi-supervised state-of-the-art algorithms by 12.2\% (1-shot) and 2.0\% (5-shots), and it is on average 0.12\% lower in the 20-shot case. 

So far, we have looked at \sys assuming that we have access to backbones pre-trained on all the auxiliary data.
However, when auxiliary data is large and does not include datasets typically used for transfer learning, it is unrealistic to pre-train backbones on all of it.
Thus, we investigate \sys using ResNet-50 pre-trained on ImageNet-1k as backbone.
In this scenario, results show that \sys outperforms distilled BiT by 5.4 points in the 1-shot setting, demonstrating its capability to exploit auxiliary data, even when they are not used to train the initial backbone.
In the 5-shot and 20-shot cases, \sys performs worse than distilled BiT (-1.2 and -6.2, respectively), but the drop is still modest considering the small amount of auxiliary data we actually use.
We also see that \sys usually outperforms all baselines with ResNet-50 backbones, often by large margins.
In cases where it does not, it is among the best performers with one exception: MetaPseudoLabel has a higher score on the 20-shot Flickr Material Dataset task.

\subsubsection{The Benefits of SCADS}\label{sec:pruning_accuracy}

In this section we study the importance of selecting task-related auxiliary data, and its influence over the performance of individual modules.
In~\Cref{tab:acc:officehome} and \Cref{tab:acc:gs_and_fms}, the performance of \sys when \scads is pruned at levels 0 and 1 (\Cref{sec:sys_spec}) offers several insights. 
First, transferring knowledge from closely related data improves \sys's performance, indeed pruning \scads causes drops in accuracy. 
Second, the performance at prune levels 0 and 1 are still often as good or better than the baselines with ResNet-50 backbone, showing that \sys can take advantage of auxiliary data even when it is more distantly related.

\begin{figure}[t]
    \centering
    \includegraphics[width=\linewidth]{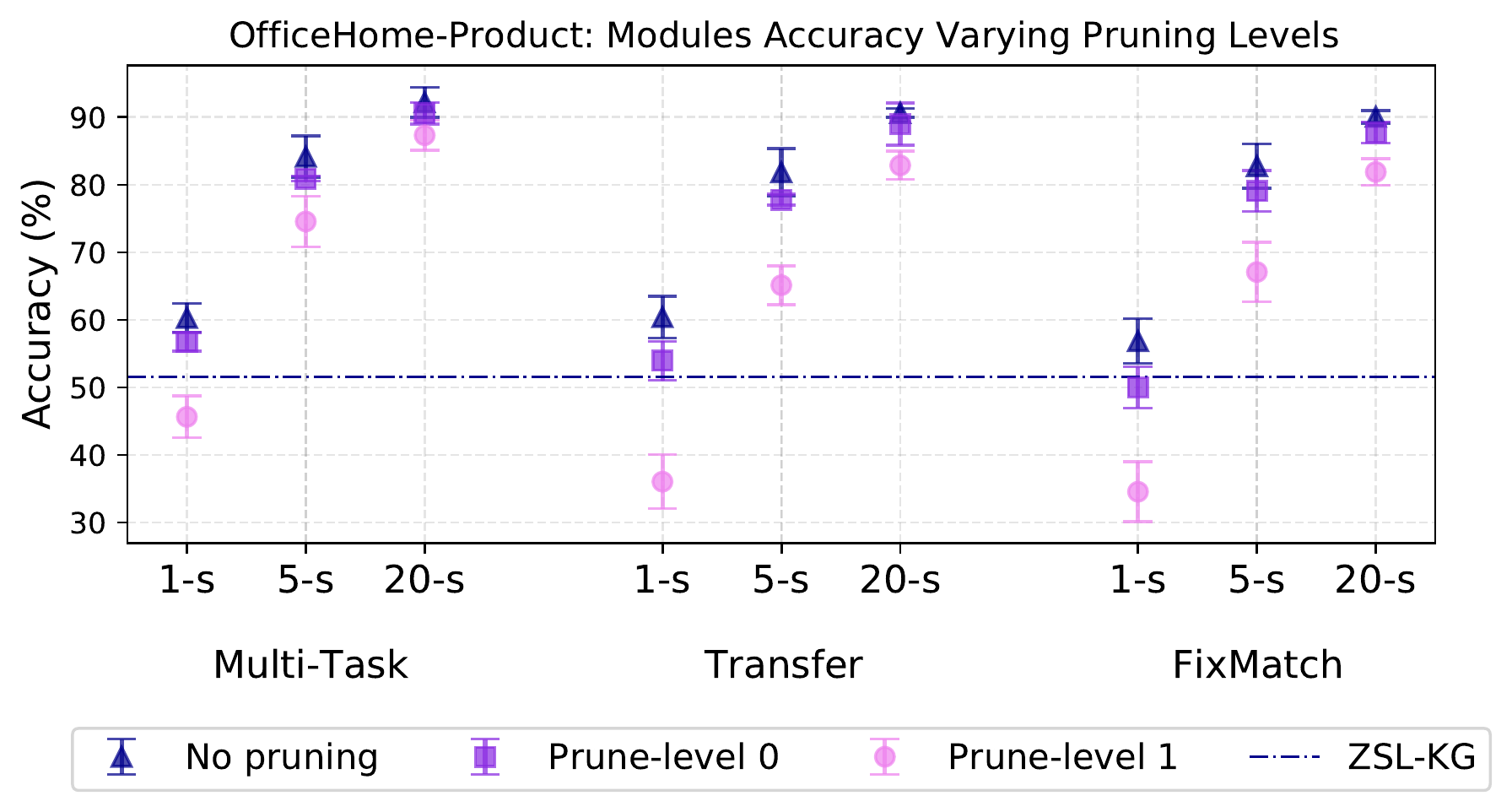}
    \vspace{0em}
    \caption{The plot shows the \%-accuracy of each module at different levels of pruning (shapes) and for varying amount of labeled examples (x-axis), on the \emph{OfficeHome-Product} dataset. The error bars indicate 95\% confidence intervals on three seeds. All modules have ResNet-50 backbone. The performances of the ZSL-KG module are invariant since it is not re-trained.}
    \label{fig:modules_pruning}
\end{figure}

For a more in-depth understanding of the importance of selected auxiliary data, we investigate the performance of the training modules at different levels of pruning. 
\Cref{fig:modules_pruning} shows that the modules benefit individually  from the use of task-related auxiliary data. 
However, we see diminishing gains when the number of labeled data increases.
Also, the Transfer and FixMatch modules significantly outperform their baseline counterparts in the no-pruning and 0-level pruning (and often in the 1-level pruning too), showing that our modifications to exploit auxiliary data are beneficial.

\subsubsection{The Benefits of Ensembling}\label{sec:intermediate_vs_final}

\begin{figure}[t]
    \centering
    \includegraphics[width=\linewidth]{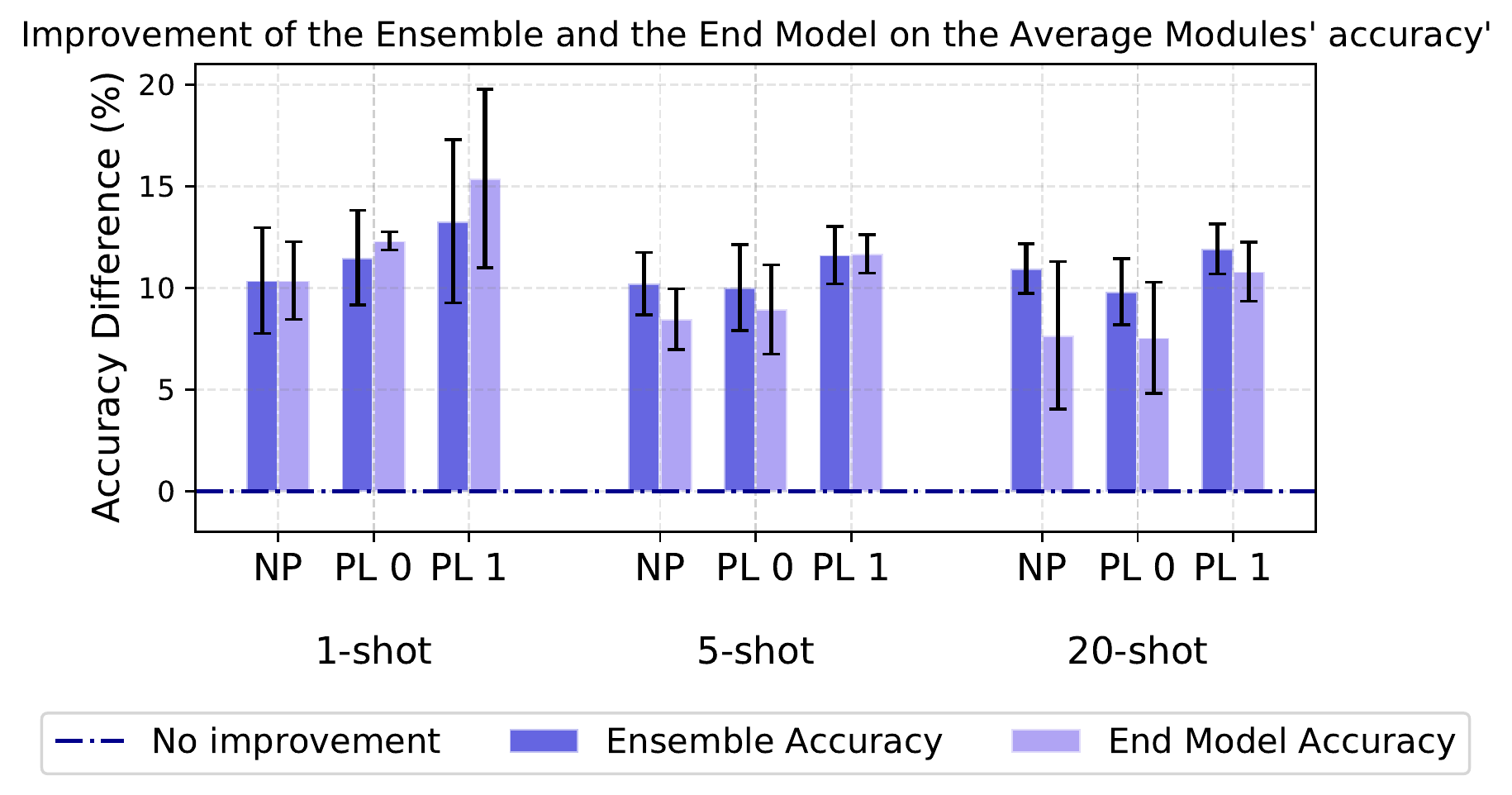}
    \vspace{0em}
    \caption{For the \emph{OfficeHome-Product} dataset, we fix the number of labeled examples (1, 5, and 20 shots) and plot the improvements of the ensemble and the end model on the average accuracy of the training modules, for different pruning levels. The results refer to \sys with ResNet-50 backbone and the 95\% confidence intervals are computed over three different seeds.}
    \label{fig:ensemble_effect}
\end{figure}

In~\Cref{fig:ensemble_effect}, we quantify the benefit of combining multiple learning techniques.
We observe that the system benefits from the modules ensemble for all the considered scenarios (i.e., varying labeled examples and pruning levels).
There is always an improvement of at least 7\% on the average accuracy of the training modules. 
We note that, in the 1-shot and 5-shots cases, the ensemble boosts the accuracy of the best performing module by at least 3.8 and 1.1 percentage points, respectively.
This suggest that the ensemble helps to boost the system performance by exploiting modules' diversity.
Additionally, we see that the efficacy of the ensembling is not corrupted when the auxiliary data is not closely related to the target classes, i.e., for different levels of pruning. 
This indicates that the ensembling procedure is crucial to limit the influence of irrelevant auxiliary data in \sys.

Finally, we investigate how the accuracy of the end model and the ensemble differ.
Aggregated results over splits and datasets show that the change in accuracy between the ensemble and the end model is variable.
Indeed, the end model performances range between -5 and 4 points on the ensemble accuracy (average -0.22). 
Overall, we believe that the occasional loss in accuracy is a fair price to pay to make more compact classifiers.
In~\Cref{app:additional_results}, we gather the plots highlighting the difference of the performances before and after the end model for each dataset and split.

\begin{figure}[t]
    \centering
    \includegraphics[width=\linewidth]{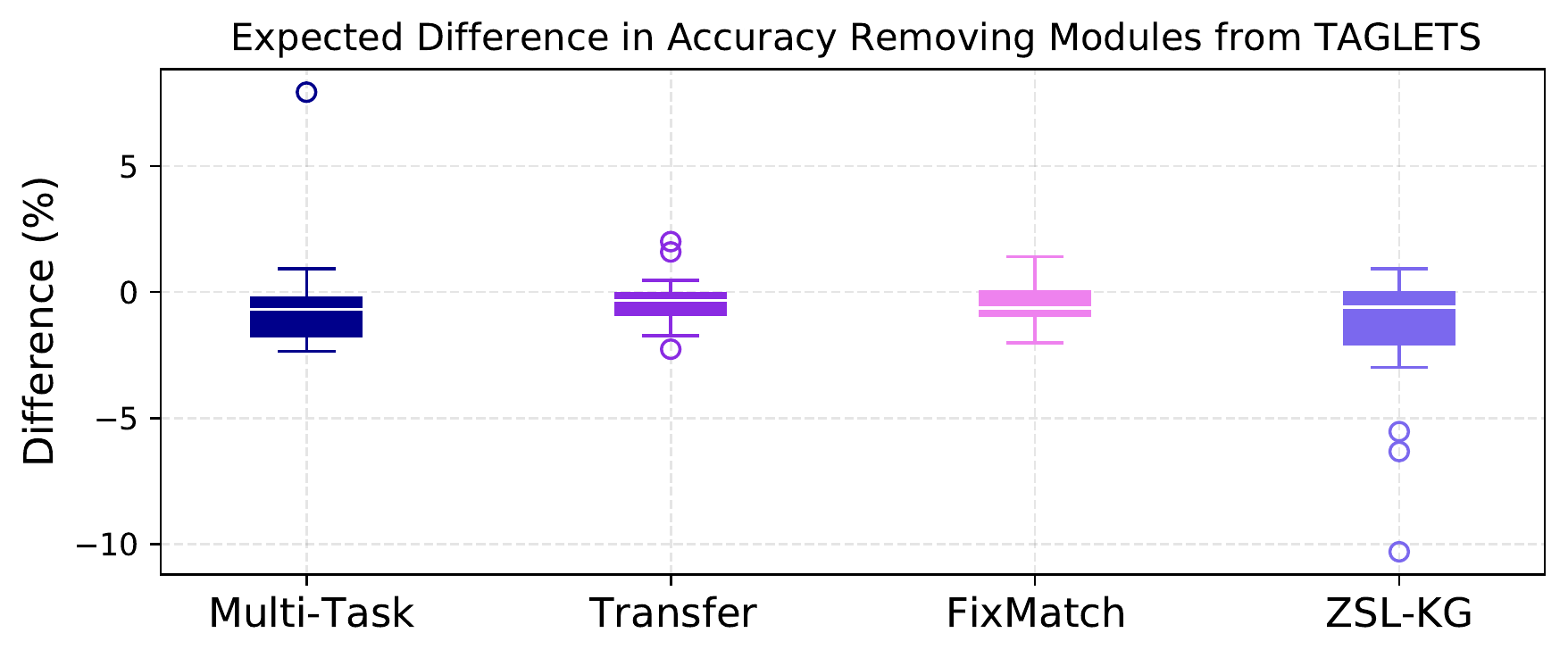}
    \vspace{0em}
    \caption{Distributions of average differences in accuracy when a module is removed from \sys. Negative values indicate that removing the module deteriorates the overall accuracy of the system. The distribution includes observations on all datasets and backbones in the 1 and 5-shots settings (\texttt{split 0}).}
    \label{fig:modules_drop}
\end{figure}

We run additional ablations to understand if each module contributes to the overall performances of \sys. 
\Cref{fig:modules_drop} shows that cutting any module reduces accuracy in at least half of the considered cases.
This suggests that each module injects diversity into the ensemble that helps the system to be robust across a large variety of settings.

\section{Conclusion}\label{sec:conclusion}

We introduce \sys, a novel system for semi-supervised learning with auxiliary data, under the limited labeled data regime. 
\sys integrates auxiliary and unlabeled data into multiple learning techniques, and it distills their knowledge into one servable model.
Our extensive experiments on datasets designed for real applications show that \sys significantly outperforms state-of-the-art algorithms in transfer and semi-supervised learning.

In this work, we emphasized the importance of exploiting auxiliary data and introduced \scads to organize it. 
However, at this stage, the component is agnostic to the data domain, which can potentially be an obstacle for learning.
This limitation leaves the door open for future works that can address the issue of multiple domains using \scads.

\section*{Acknowledgements}
We thank Michael Littman, Eli Upfal, and James Tompkin for many helpful and insightful discussions throughout the design and development of TAGLETS.
This material is based on research sponsored by Defense Advanced Research Projects Agency (DARPA) and Air Force Research Laboratory (AFRL) under agreement number FA8750-19-2-1006. The U.S. Government is authorized to reproduce and distribute reprints for Governmental purposes notwithstanding any copyright notation thereon. The views and conclusions contained herein
are those of the authors and should not be interpreted as necessarily representing the official policies or endorsements, either expressed or implied, of Defense Advanced Research Projects Agency (DARPA) and Air Force Research Laboratory (AFRL) or the U.S. Government. We gratefully acknowledge support from Google and Cisco. Disclosure: Stephen Bach is an advisor to Snorkel AI, a company that
provides software and services for weakly supervised machine learning.

\nocite{langley00}

\bibliography{mlsys}
\bibliographystyle{mlsys2022}

\clearpage
\appendix
\section{Appendix}

\subsection{SCADS Embeddings details}\label{app:scads}
We learn  \emph{SCADS embeddings} using the \emph{expanded retrofitting technique} introduced by~\citep{faruqui:retrofitting,speer:ensemble}, which expresses both the network topology (i.e., relations among concepts) and word embeddings learned from text, i.e., word2vec~\cite{mikolov:word2veca,mikolov:word2vecb}.
In a nutshell, for each concept $\concept$, the method learns an embedding $\scadsemb{\concept}$ such that it is close to the original word vector $\wordemb{\concept}$ and to the neighborhood $\neigh$ of $\concept$ in the graph. 
Formally, the objective function $\objemb{\setconcept}$ to minimize is: 
\begin{equation}\label{eq:scadsembedding}
\objemb{\setconcept} = \sum_{i=i}^n\left [ \alpha_i \left \| \wordemb{i} - \scadsemb{i} \right \|^2 + \sum_{(i,j) \in \neigh} \beta_{ij}\left \| \scadsemb{i} - \wordemb{j} \right \|^2 \right ] 
\end{equation}
where $\alpha_i$ indicates the importance of the old word embedding, and $\beta_{ij}$ is the edge weight.
Following \cite{speer:conceptnet}, we set $\alpha=0$ to handle out-of-vocabulary concepts.

\subsection{Experimental setting details}\label{app:experiments}
As explained in ~\Cref{sec:sys_spec}, for each of the experimental datasets, except Grocery Store Dataset since it already has a pre-determined test set, we first partition the dataset into a train-test split by randomly take a fixed number of images per class from the pool of data to use as the test set and assign the rest for the train set. 
The number of test images per class is 5, 10 and 10 for FMD and OfficeHome-Product, and OfficeHome-Clipart respectively. 
Next, we randomly choose a fixed number of images per class (1, 5, or 20) in the train partition to be labeled and use the rest of the train images as unlabeled images. 
We do not report the 20-shot results of Grocery Store Dataset because the minimum number of images per class for this dataset is 18. 
We use the same seed for both partitioning a dataset into train-test split and subsequently choosing train images in the dataset to be labeled, and we report the results of three different seeds in~\Cref{tab:acc:gs_and_fms_split1,tab:acc:officehome_split1}. 
The results are reported as 95\% confidence interval over 3 different training seeds, which affects the initialization of linear layers we append to the backbones and the data shuffling during training).

\subsection{Training details}\label{app:training}
For data augmentation during training, we apply random resized crop, where the cropping scale is randomly chosen between 0.8 and 1, and random horizontal flip.
All images are resized to 224x224 and normalized with ImageNet1k's statistics, both are community standards. 
For all modules, except ZSL-KG since it is a zero-shot module, the classifier head(s) appended to ResNet-50 (ImageNet-1k) is a single fully-connected layer, while following \cite{kolesnikov:bigtransfer}, the classifier head(s) appended to BiT (ImageNet-21k) is a 2-D convolutional layer with kernel size equal to 1x1.
The training details of individual modules follow.

\textbf{Transfer Module}.
The batch size is set to 256. The optimizer is SGD with a learning rate of 0.003 and a momentum of 0.9. 
When using ResNet-50 (ImageNet-1k), we first fine-tune the model on the selected auxiliary data for 5 epochs, and then we fine-tune the model on the target data for 40 epochs with the learning rate decayed by 0.1 at epoch 20 and 30.  
When using BiT (ImageNet-21k), we fine-tune the model on the selected auxiliary data for 2000 steps, and then we fine-tune the model on the target data for 500 steps. 
Linear learning rate warmup is used for the first 100 and 500 steps when the training steps are 500 and 2000 respectively, and then we decay the learning rate at step (200, 300, 400) and (900, 1300, 1700, 2000) by 0.1 when the training steps are 500 and 2000 respectively.

\begin{table*}[t]
    \centering
    \resizebox{\linewidth}{!}{
    \begin{tabular}{lccccccc}
    \toprule
    \multirow{2}{*}{Method} & \multirow{2}{*}{Backbone} & \multicolumn{3}{c}{OfficeHome-Product} & \multicolumn{3}{c}{OfficeHome-Clipart}\\
    \cmidrule(lr){3-8}
     &  & 1-shot & 5-shot & 20-shot & 1-shot & 5-shot & 20-shot\\    
    \midrule
    \textbf{\texttt{Split 1}} & &  &  &  &  &  & \\
    \midrule
    Fine-tuning & BiT (ImageNet-21k) & 62.56 $\pm$ 2.10 & 84.72 $\pm$ 0.88 & 91.85 $\pm$ 0.38 & 31.85 $\pm$ 4.63 & 65.95 $\pm$ 0.58 & \textbf{80.67 $\pm$ 0.58}\\
    Fine-tuning (Distilled) & BiT (ImageNet-21k) & 62.56 $\pm$ 2.54 & \textbf{85.74 $\pm$ 0.58} & \textbf{92.51 $\pm$ 0.44} & 31.18 $\pm$ 2.46 & 65.44 $\pm$ 0.96 & 78.51 $\pm$ 1.89\\
    FixMatch & BiT (ImageNet-21k) & 56.87 $\pm$ 3.11 & 83.95 $\pm$ 0.96 & 91.64 $\pm$ 1.17 & 28.87 $\pm$ 9.40 & 64.56 $\pm$ 0.58 & 81.59 $\pm$ 1.23\\
    Meta Pseudo Label & BiT (ImageNet-21k) & 61.28 $\pm$ 1.12 & 80.97 $\pm$ 2.60 & 90.77 $\pm$ 1.38 & 31.64 $\pm$ 0.96 & 62.67 $\pm$ 1.72 & 76.15 $\pm$ 1.01\\
    TAGLETS & BiT (ImageNet-21k) &\textbf{ 74.67 $\pm$ 1.45} & \textbf{85.85 $\pm$ 0.66} & 91.95 $\pm$ 0.88 & \textbf{43.64 $\pm$ 5.52} & \textbf{68.05 $\pm$ 1.45} & \textbf{80.41 $\pm$ 3.67}\\
    \midrule
    Fine-tuning & ResNet-50 (ImageNet-1k) & 34.72 $\pm$ 1.10 & 64.97 $\pm$ 0.88 & 84.41 $\pm$ 2.49 & 17.03 $\pm$ 1.89 & 43.85 $\pm$ 3.27 & 69.64 $\pm$ 3.11\\
    Fine-tuning (Distilled) & ResNet-50 (ImageNet-1k) & 35.95 $\pm$ 2.46 & 65.44 $\pm$ 3.85 & 85.79 $\pm$ 0.58 & 16.72 $\pm$ 1.81 & 45.13 $\pm$ 2.49 & 70.92 $\pm$ 2.13\\
    FixMatch & ResNet-50 (ImageNet-1k) & 43.74 $\pm$ 0.80 & 74.31 $\pm$ 1.67 & 85.90 $\pm$ 0.58 & 22.51 $\pm$ 1.17 & 51.38 $\pm$ 3.03 & 67.03 $\pm$ 0.22\\
    Meta Pseudo Label & ResNet-50 (ImageNet-1k) & 51.69 $\pm$ 2.89 & 76.76 $\pm$ 1.38 & 89.75 $\pm$ 0.88 & 23.48 $\pm$ 1.59 & 56.10 $\pm$ 3.18 & \underline{75.40 $\pm$ 1.96}\\
    TAGLETS & ResNet-50 (ImageNet-1k) & \underline{70.92 $\pm$ 3.13} & \underline{82.67 $\pm$ 1.10} & \underline{90.72 $\pm$ 0.96} & \underline{43.28 $\pm$ 0.96} & \underline{64.77 $\pm$ 2.51} & \underline{75.64 $\pm$ 2.68}\\
    \midrule
    TAGLETS prune-level 0 & ResNet-50 (ImageNet-1k) & \underline{68.87 $\pm$ 2.46} & \underline{81.59 $\pm$ 0.58} & \underline{90.51 $\pm$ 1.81} & \underline{42.82 $\pm$ 1.45} & 60.10 $\pm$ 2.10 & \underline{76.05 $\pm$ 2.17}\\
    TAGLETS prune-level 1 & ResNet-50 (ImageNet-1k) & 64.00 $\pm$ 3.40 & 77.59 $\pm$ 1.23 & 88.67 $\pm$ 0.80 & 37.69 $\pm$ 2.29 & 53.23 $\pm$ 2.39 & \underline{74.00 $\pm$ 1.75}\\
    \midrule
    \midrule
    \textbf{\texttt{Split 2}} & &  &  &  &  &  & \\
    \midrule
    Fine-tuning & BiT (ImageNet-21k) & 57.49 $\pm$ 1.54 & \textbf{82.51 $\pm$ 1.23} & \textbf{90.56 $\pm$ 0.58} & 32.87 $\pm$ 3.43 & 62.87 $\pm$ 1.10 & \textbf{78.15 $\pm$ 1.32}\\
    Fine-tuning (Distilled) & BiT (ImageNet-21k) & 56.62 $\pm$ 3.99 & \textbf{82.77 $\pm$ 1.75} & \textbf{90.77 $\pm$ 0.66} & 32.92 $\pm$ 5.88 & 62.87 $\pm$ 2.10 & \textbf{78.10 $\pm$ 0.44}\\
    FixMatch & BiT (ImageNet-21k) & 48.15 $\pm$ 1.91 & \textbf{82.41 $\pm$ 1.81} & \textbf{90.46 $\pm$ 0.76} & 9.59 $\pm$ 6.09 & 63.23 $\pm$ 2.65 & \textbf{79.23 $\pm$ 1.38}\\
    Meta Pseudo Label & BiT (ImageNet-21k) & 57.79 $\pm$ 0.58 & 78.61 $\pm$ 1.52 & 88.21 $\pm$ 0.58 & 26.46 $\pm$ 3.90 & 54.31 $\pm$ 2.12 & 75.12 $\pm$ 0.79\\
    TAGLETS & BiT (ImageNet-21k) & \textbf{70.00 $\pm$ 0.76} & \textbf{84.67 $\pm$ 2.30} & \textbf{90.56 $\pm$ 1.72} & \textbf{45.79 $\pm$ 4.19} & \textbf{66.62 $\pm$ 2.13} & \textbf{79.08 $\pm$ 1.32}\\
    \midrule 
    Fine-tuning & ResNet-50 (ImageNet-1k) & 34.36 $\pm$ 3.45 & 63.54 $\pm$ 3.50 & 82.10 $\pm$ 0.96 & 18.92 $\pm$ 1.99 & 41.69 $\pm$ 2.39 & 70.72 $\pm$ 3.55\\
    Fine-tuning (Distilled) & ResNet-50 (ImageNet-1k) & 35.49 $\pm$ 3.83 & 65.90 $\pm$ 3.25 & 84.62 $\pm$ 0.66 & 20.26 $\pm$ 0.44 & 43.85 $\pm$ 4.63 & 72.10 $\pm$ 2.97\\
    FixMatch & ResNet-50 (ImageNet-1k) & 43.38 $\pm$ 3.65 & 72.15 $\pm$ 0.76 & 84.56 $\pm$ 2.10 & 24.05 $\pm$ 1.81 & 49.38 $\pm$ 3.40 & 69.85 $\pm$ 2.39\\
    Meta Pseudo Label & ResNet-50 (ImageNet-1k) & 50.87 $\pm$ 2.10 & 74.87 $\pm$ 2.76 & 88.20 $\pm$ 1.45 & 21.28 $\pm$ 1.22 & 44.41 $\pm$ 1.45 & 70.41 $\pm$ 2.10\\
    TAGLETS & ResNet-50 (ImageNet-1k) & \underline{67.54 $\pm$ 0.00} & \underline{84.62 $\pm$ 1.38} & \underline{89.90 $\pm$ 1.23} & \underline{43.95 $\pm$ 2.10} & \underline{63.18 $\pm$ 1.45} & \underline{78.31 $\pm$ 3.03}\\
    \midrule
    TAGLETS prune-level 0 & ResNet-50 (ImageNet-1k) & 65.33 $\pm$ 0.58 & 82.67 $\pm$ 1.92 & \underline{89.64 $\pm$ 0.22} & \underline{43.59 $\pm$ 1.96} & 59.33 $\pm$ 1.10 & \underline{77.18 $\pm$ 2.54}\\
    TAGLETS prune-level 1 & ResNet-50 (ImageNet-1k) & 60.56 $\pm$ 1.59 & 78.00 $\pm$ 1.53 & \underline{88.82 $\pm$ 0.80} & 39.59 $\pm$ 3.45 & 55.08 $\pm$ 1.99 & \underline{75.49 $\pm$ 1.17}\\
    \bottomrule
    \end{tabular}}
    \caption{Accuracy (\%) of \sys and baselines on \emph{OfficeHome Product} and \emph{Clipart} (\texttt{Split 1 and 2}) for increasing number of labeled examples (1, 5, and 20 shots), different backbones (ResNet-50 trained on ImageNet-1k and BiT trained on ImageNet-21k) and levels of pruning (no-pruning, 0 and 1). We make bold and underline the best performances, and all those within their 95\% confidence intervals, carried out using \textbf{BiT} and \underline{ResNet} backbones, respectively. Measurements are repeated for three seeds and we report the 95\% confidence intervals.}
\label{tab:acc:officehome_split1}
\end{table*}
\begin{table*}[ht]
    \centering
    \resizebox{\linewidth}{!}{
    \begin{tabular}{lcccccc}
    \toprule
    \multirow{2}{*}{Method} & \multirow{2}{*}{Backbone} & \multicolumn{2}{c}{Grocery Store Dataset} & \multicolumn{3}{c}{Flickr Material Dataset}\\
    \cmidrule(lr){3-7}
     &  & 1-shot & 5-shot &  1-shot & 5-shot & 20-shot\\    
    \midrule
    \textbf{\texttt{Split 1}} & &  &  &  &  & \\
    \midrule
    Fine-tuning & BiT (ImageNet-21k) & 60.52 $\pm$ 4.55 & 84.02 $\pm$ 1.06 & 42.47 $\pm$ 1.88 & 76.27 $\pm$ 3.86 & \textbf{87.40 $\pm$ 0.86}\\
    Fine-tuning (Distilled) & BiT (ImageNet-21k) & 60.56 $\pm$ 5.02 & 85.42 $\pm$ 2.61 & 45.07 $\pm$ 2.45 & \textbf{77.93 $\pm$ 0.29} & \textbf{87.67 $\pm$ 1.15}\\
    FixMatch & BiT (ImageNet-21k) & 48.58 $\pm$ 13.41 & \textbf{87.61 $\pm$ 4.82} & 12.60 $\pm$ 11.19 & 70.73 $\pm$ 10.84 & 86.47 $\pm$ 2.07\\
    Meta Pseudo Label & BiT (ImageNet-21k) & 59.52 $\pm$ 3.96 & 85.58 $\pm$ 0.75 & 44.73 $\pm$ 5.35  & 75.20 $\pm$ 3.44 & 80.73 $\pm$ 2.45\\
    TAGLETS & BiT (ImageNet-21k) & \textbf{66.26 $\pm$ 5.60} & \textbf{87.87 $\pm$ 0.95} & \textbf{60.80 $\pm$ 3.97} &77.60 $\pm$ 4.24 & \textbf{86.67 $\pm$ 1.03}\\
    \midrule
    Fine-tuning & ResNet-50 (ImageNet-1k) & 34.86 $\pm$ 4.11 & 67.23 $\pm$ 0.80 & 25.33 $\pm$ 2.83 & 41.53 $\pm$ 2.91 & 61.93 $\pm$ 3.86\\
    Fine-tuning (Distilled) & ResNet-50 (ImageNet-1k) & 36.31 $\pm$ 1.70 & 72.22 $\pm$ 2.91 & 26.80 $\pm$ 1.31 & 40.80 $\pm$ 6.88 & 64.80 $\pm$ 3.48\\
    FixMatch & ResNet-50 (ImageNet-1k) & 32.69 $\pm$ 1.85 & 71.60 $\pm$ 4.17 & 28.80 $\pm$ 4.24 & 58.80 $\pm$ 4.74 & 75.20 $\pm$ 1.31\\
    Meta Pseudo Label & ResNet-50 (ImageNet-1k) & 38.19 $\pm$ 5.33 & 75.51 $\pm$ 0.47 & 38.44 $\pm$ 10.20 & 65.13 $\pm$ 5.35 & \underline{78.07 $\pm$ 2.50}\\
    TAGLETS & ResNet-50 (ImageNet-1k) & \underline{56.45 $\pm$ 1.90} & \underline{82.43 $\pm$ 1.76} & \underline{51.87 $\pm$ 3.38} & \underline{68.53 $\pm$ 2.24 }& 73.53 $\pm$ 1.43\\
    \midrule
    TAGLETS prune-level 0 & ResNet-50 (ImageNet-1k) & 49.99 $\pm$ 1.85 & 80.24 $\pm$ 2.07 & \underline{48.53 $\pm$ 2.29} & 62.20 $\pm$ 4.74 & 73.00 $\pm$ 2.28\\
    TAGLETS prune-level 1 & ResNet-50 (ImageNet-1k) & 42.91 $\pm$ 9.81 & 73.35 $\pm$ 3.25 & \underline{50.07 $\pm$ 5.80} & \underline{66.67 $\pm$ 3.76} & 72.93 $\pm$ 1.03\\
    \midrule
    \midrule
    \textbf{\textbf{\texttt{Split 2}}} & & &  &  & &  \\
    \midrule
    Fine-tuning & BiT (ImageNet-21k) & 60.08 $\pm$ 1.66 & 83.61 $\pm$ 0.71 & 41.07 $\pm$ 6.96 & 69.40 $\pm$ 2.17 & \textbf{85.07 $\pm$ 1.52}\\
    Fine-tuning (Distilled) & BiT (ImageNet-21k) & 62.72 $\pm$ 2.66 & \textbf{84.10 $\pm$ 1.14} & 42.53 $\pm$ 6.21 & 70.93 $\pm$ 2.74 & \textbf{84.53 $\pm$ 1.03}\\
    FixMatch & BiT (ImageNet-21k) & 60.16 $\pm$ 6.51 & \textbf{87.08 $\pm$ 3.30} & 12.13 $\pm$ 8.32 &\textbf{ 72.40 $\pm$ 1.31} & \textbf{83.60 $\pm$ 2.77}\\
    Meta Pseudo Label & BiT (ImageNet-21k) & 62.12 $\pm$ 3.12 & 82.19 $\pm$ 0.21 & 48.86 $\pm$ 8.90 &\textbf{ 74.06 $\pm$ 2.01} & 78.79 $\pm$ 1.31\\
    TAGLETS & BiT (ImageNet-21k) & \textbf{68.26 $\pm$ 3.29} & \textbf{85.88 $\pm$ 0.56} & \textbf{59.60 $\pm$ 9.48} & \textbf{73.47 $\pm$ 2.87} & \textbf{84.27 $\pm$ 1.88}\\
    \midrule
    Fine-tuning & ResNet-50 (ImageNet-1k) & 31.99 $\pm$ 5.14 & 65.78 $\pm$ 4.85 & 28.00 $\pm$ 4.07 & 44.40 $\pm$ 1.72 & 58.80 $\pm$ 5.16\\
    Fine-tuning (Distilled) & ResNet-50 (ImageNet-1k) & 33.17 $\pm$ 2.66 & 69.59 $\pm$ 3.50 & 27.60 $\pm$ 10.57 & 46.07 $\pm$ 7.03 & 62.73 $\pm$ 5.27\\
    FixMatch & ResNet-50 (ImageNet-1k) & 37.64 $\pm$ 1.50 & 68.10 $\pm$ 2.86 & 35.80 $\pm$ 12.41 & 57.40 $\pm$ 5.23 & 69.73 $\pm$ 2.55\\
    Meta Pseudo Label & ResNet-50 (ImageNet-1k) & 39.46 $\pm$ 0.97 & 70.03 $\pm$ 1.35 & 38.66 $\pm$ 2.57 & \underline{65.66 $\pm$ 5.31} & \underline{75.46 $\pm$ 2.99}\\
    TAGLETS & ResNet-50 (ImageNet-1k) & \underline{57.45 $\pm$ 5.66} & \underline{81.57 $\pm$ 1.04} & \underline{53.13 $\pm$ 6.29} & \underline{65.87 $\pm$ 1.74} & \underline{72.80 $\pm$ 2.28}\\
    \midrule
    TAGLETS prune-level 0 & ResNet-50 (ImageNet-1k) & \underline{55.36 $\pm$ 6.00} & 79.38 $\pm$ 2.47 & \underline{51.67 $\pm$ 3.31} & \underline{64.53 $\pm$ 2.99} & 71.80 $\pm$ 3.48\\
    TAGLETS prune-level 1 & ResNet-50 (ImageNet-1k) & 47.03 $\pm$ 8.44 & 74.98 $\pm$ 1.36 & 46.00 $\pm$ 15.30 & 62.00 $\pm$ 7.81 & 69.93 $\pm$ 2.24\\
    \bottomrule
     \end{tabular}}
    \caption{Accuracy (\%) of \sys and baselines on \emph{Grocery Store} and \emph{Flicker Material Datasets} (\texttt{Split 1 and 2}) for increasing number of labeled examples (1, 5, and 20 shots), different backbones (ResNet-50 trained on ImageNet-1k and BiT trained on ImageNet-21k) and levels of pruning (no-pruning, 0 and 1). We make bold and underline the best performances, and all those within their 95\% confidence intervals, carried out using \textbf{BiT} and \underline{ResNet} backbones, respectively. Measurements are repeated for three seeds and we report the 95\% confidence intervals.}
\label{tab:acc:gs_and_fms_split1}
\end{table*}

\textbf{Multi-task Module}.
The batch size is set to 128. The optimizer is SGD with a learning rate of 0.003 and a momentum of 0.9. 
When using ResNet-50 (ImageNet-1k), we train the model for 8 epochs conditioned on the selected auxiliary data, and we decay the learning rate by 0.1 at epoch 4 and 6. 
When using BiT (ImageNet-21k), we train the model for 2000 steps. 
Linear learning rate warmup is used for the first 500 steps, and then we decay the learning rate at step 900, 1300, and 1700 by 0.1. 

\textbf{FixMatch Module}.
We first pretrain the ResNet-50 (ImageNet-1k) or BiT (ImageNet-21k) backbone on selected auxiliary data for five epochs using an SGD optimizer. 
During pretraining, we set the learning rate to 0.003, momentum is set to 0.9, and the batch size is set to 256. 
After pretraining, we follow the training procedures outlined in \cite{sohn:fixmatch}. 
We use an SGD optimizer with Nesterov momentum. The optimizer's learning rate is set to 0.0005, and its momentum is set to 0.9.
The model is trained  on labeled and unlabeled data for 30 epochs when using ResNet-50 (ImageNet-1k) and 500 epochs when using BiT (ImageNet-21k). 
Both backbones use a batch size of 128. 
After each batch, the learning rate is set according to the cosine learning decay rate $\eta\cos \left( \frac{7\pi k}{16K}\right)$, where $\eta$ is the initial learning rate, $k$ is the current batch number, and $K$ is the total number of batches seen during training.

\textbf{ZSL-KG Module}.
In our experiments, we follow the object classification implementation from \citet{nayak:arxiv20} \footnote{\href{https://github.com/BatsResearch/nayak-arxiv20-code}{https://github.com/BatsResearch/nayak-arxiv20-code}}.
We pretrain ZSL-KG by minimizing the L2 distance between the learned class representations and the weights of the fully connected layer from a pretrained classifier:
\begin{equation}
    \mathcal{L}_{Z} = \frac{1}{n}\sum_{i}^{n}\left( w_{i} - z_{i}\right)^{2}
\end{equation}
where $w_{i}$ is the pretrained weight from the classifier and $z_{i}$ is the class representation from the ZSL-KG module for the $i$-th class. 
Following \citet{nayak:arxiv20}, we train ZSL-KG on the ConceptNet graph. 
To train the ZSL-KG module with the ConceptNet graph, we use the fully connected layer weights from ResNet101 pretrained on ILSVRC2012~\citep{russakovsky:ijcv15} obtained from Torchvision~\citep{marcel:icm10}.
The module is trained for 1000 epochs on a random 950/50 class split where 950 classes are used for training while the rest are used for validation.
During the pretraining, we use the Adam optimizer with a learning rate of $1e-03$ and weight decay of $5e-04$. 
The model checkpoint with the least loss on the validation classes is used in \scads.

\textbf{End Model}
The batch size is set to 256. When using ResNet-50 (ImageNet-1k), we train the model for 30 epochs using Adam optimizer with learning rate 0.0005 and weight decay 0.0001.
The learning rate is decayed by 0.1 at epoch 20. 
When using BiT (ImageNet-21k), we train the model for 500 steps. The optimizer is SGD with a learning rate of 0.003 and a momentum of 0.9. 
Linear learning rate warmup is used for the first 100 steps, and then we decay the learning rate at step 200, 300, and 400 by 0.1. 

\textbf{Meta Pseudo Labels} As mentioned in \Cref{sec:baselines}, Meta Pseudo Labels is comprised of a teacher model and a student model. The teacher has either a ResNet-50 (ImageNet-1k) or BiT (ImageNet-21k) backbone, and the student has ResNet-50 (ImageNet-1k) backbone. During training, both the student and teacher use an SGD optimizer with momentum set to 0.9. The teacher's initial learning rate is set to 0.0005, and the student's initial learning rate is set to 0.001 for all datasets except for Grocery Store, which uses a learning rate of 0.0001 to mitigate early convergence. The student and teacher are trained for 500 steps using a batch size of 128. The learning rate of both models are set using according to the cosine learning decay rate  $\frac{\eta}{2}\left( 1 + \cos \frac{\pi k}{K}\right)$, where $\eta$ is the initial learning rate, $k$ is the current batch number, and $K$ is the total number of batches seen during training. After teacher-student training, the student is fine-tuned for 30 epochs using SGD with momentum set to 0.9 and a fixed learning rate of 0.003.

\subsection{Additional results}\label{app:additional_results}

In this section, we collect the experimental results on \texttt{split 1} and \texttt{split 2} of the datasets (\Cref{sec:sys_spec}). 
\Cref{tab:acc:gs_and_fms_split1,tab:acc:officehome_split1} report the trends discussed in~\Cref{sec:accuracy}: (1) \sys outperforms the baselines in the 1 and 5 shots settings, and competitive in the 20-shots case. (2) The system is robust to the amount on auxiliary data its backbone is pre-trained., i.e., \sys with ResNet-50 (ImageNet-1k) achieves better scores than distilled fine tuning of BiT (ImageNet-21k).

\begin{figure}[t]
\centering
\includegraphics[width=0.4\textwidth]{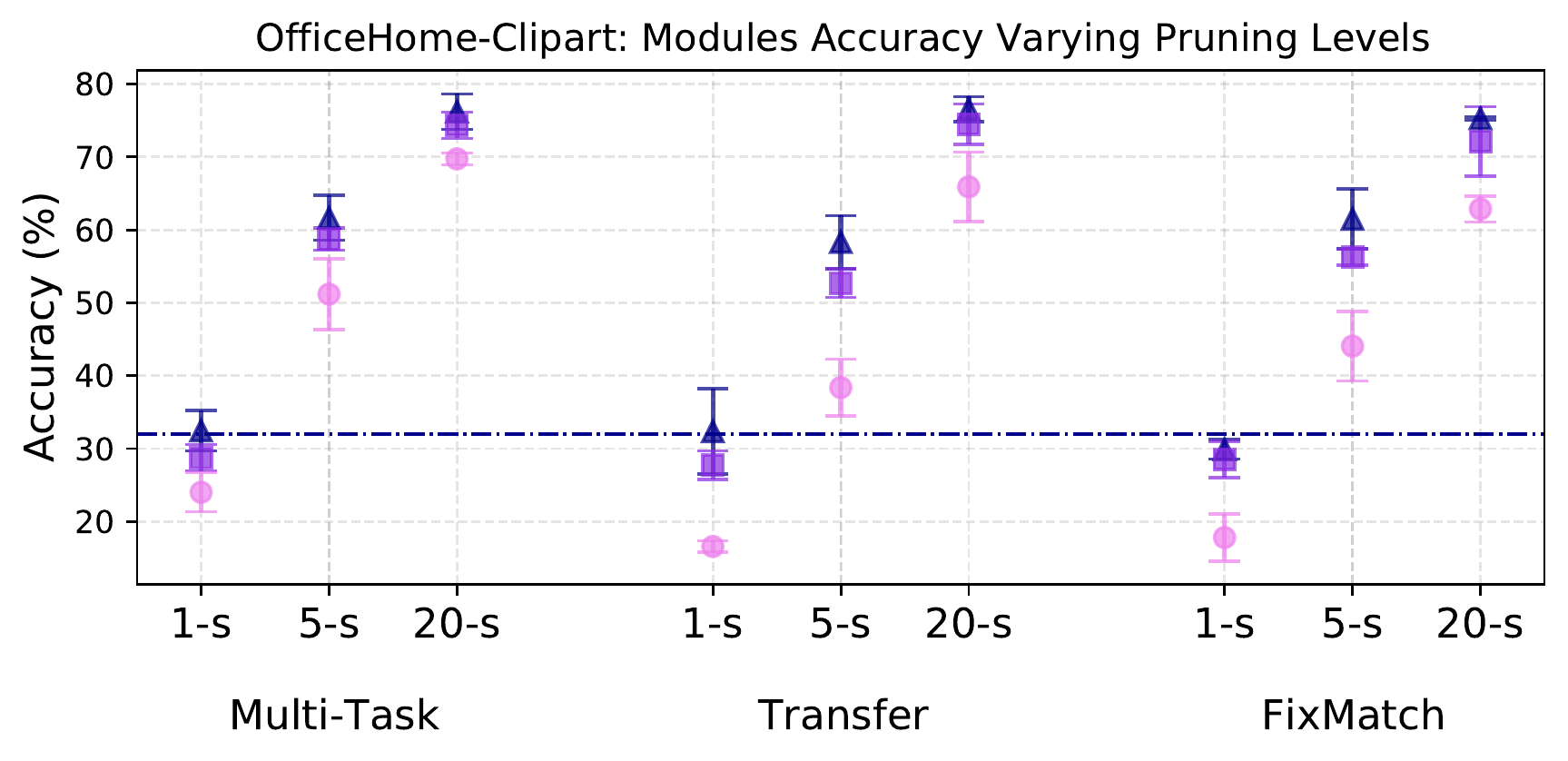}
\includegraphics[width=0.4\textwidth]{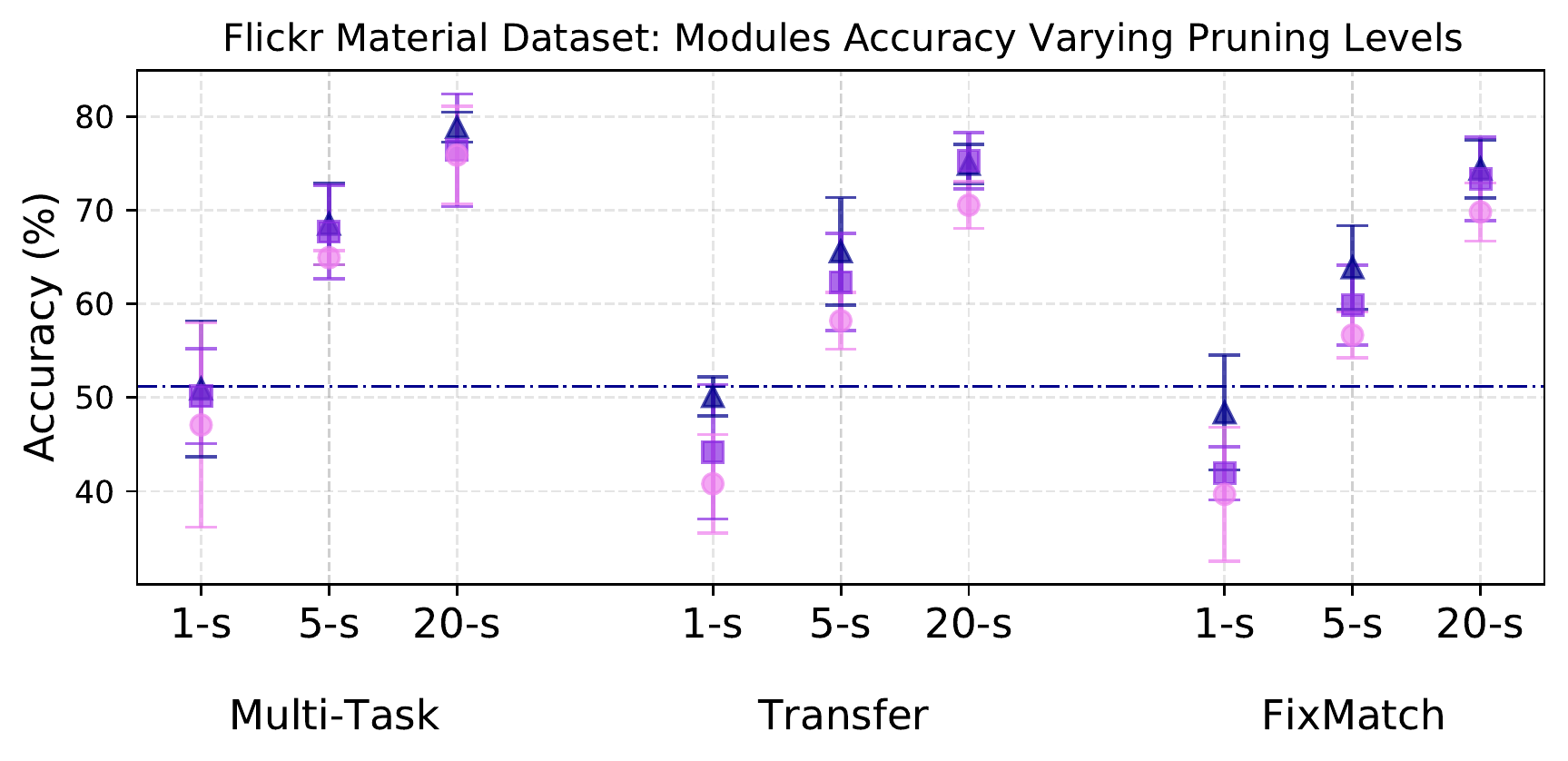}
\includegraphics[width=0.4\textwidth]{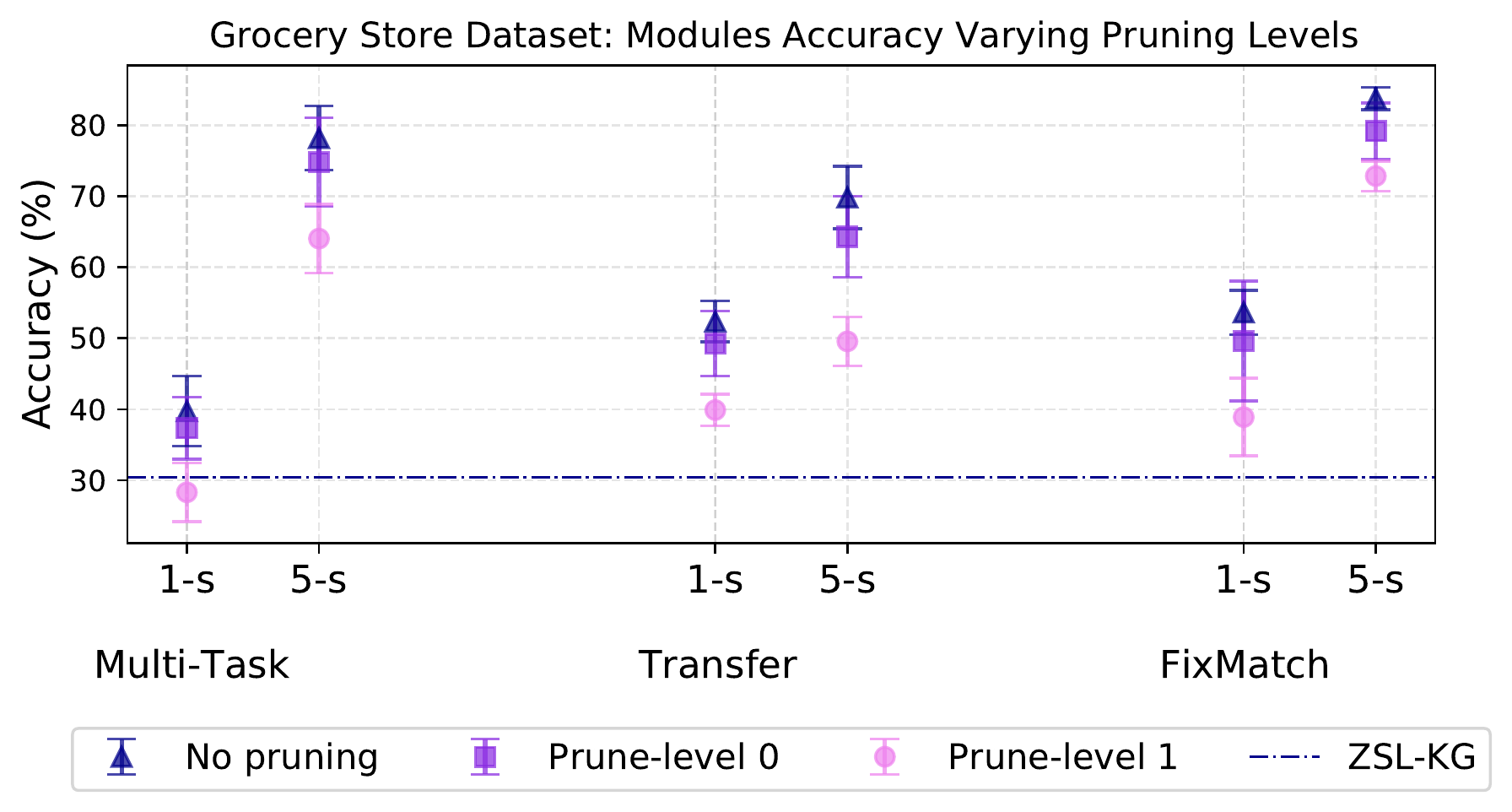}
\caption{(\texttt{Split 0}) The plot shows the \%-accuracy of each module at different levels of pruning (shapes) and for varying amount of labeled examples (x-axis), on the \emph{OfficeHome-Clipart}, \emph{Flick Material} and \emph{Grocery Store} datasets.The error bars indicate 95\% confidence intervals on three seeds. All modules have ResNet-50 backbone. We note that the performances of ZSL-KG module are invariant to pruning, since it is used as pre-trained model on ConceptNet.}
\label{fig:split_0_pruning}
\end{figure}

\Cref{fig:split_0_pruning} and the pruning results in~\Cref{tab:acc:gs_and_fms_split1,tab:acc:officehome_split1} confirm that the more the selected auxiliary data are far from the target classes, the more the accuracy of the individual modules decreases. 
Moreover, as we increase the size of labeled data, the importance of relatedness between the auxiliary data and the target classes reduces. 
Results on the other splits of the datasets suggest the same observations.

\begin{figure}[ht]
\centering
\includegraphics[width=0.4\textwidth]{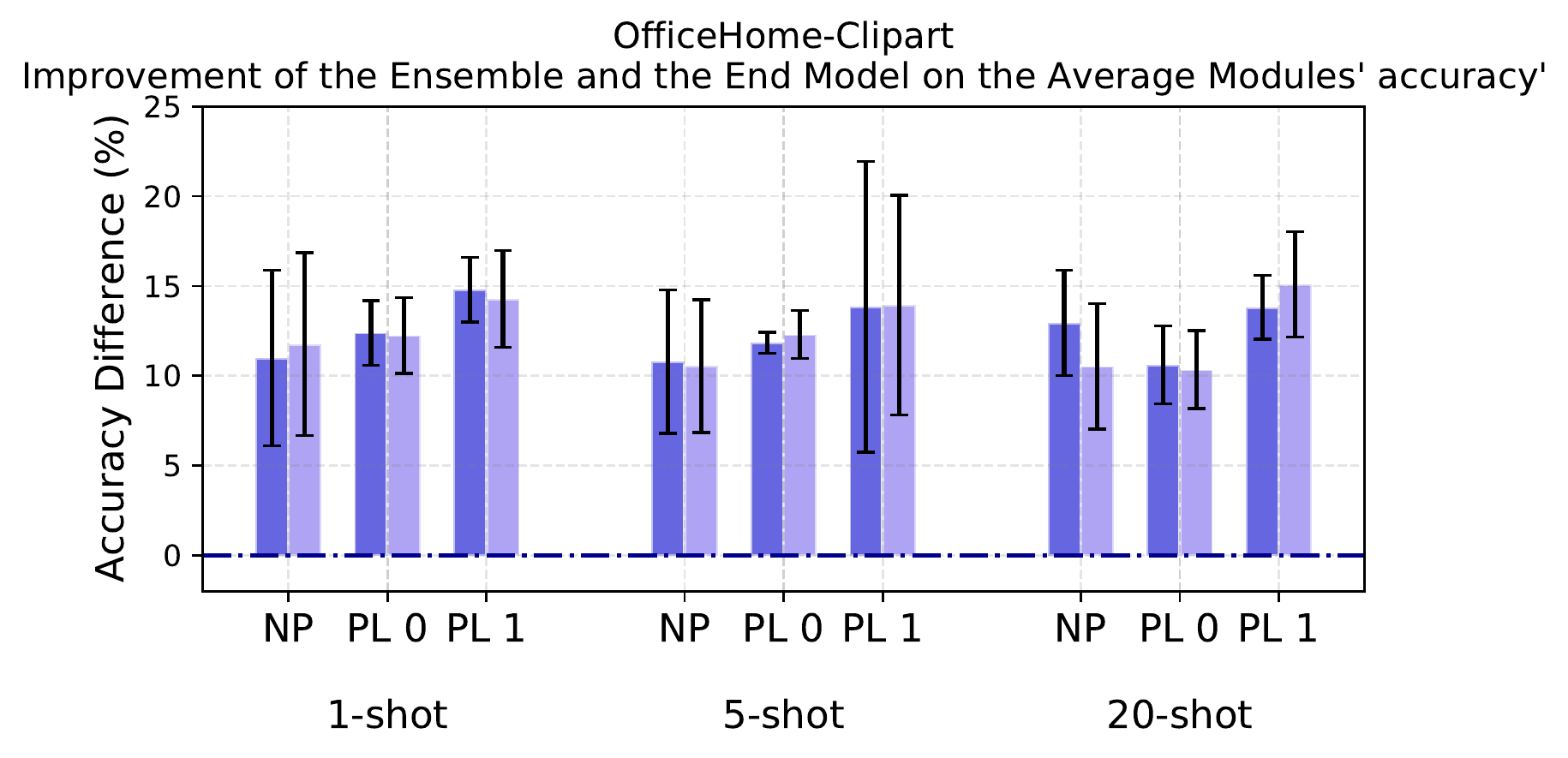}
\includegraphics[width=0.4\textwidth]{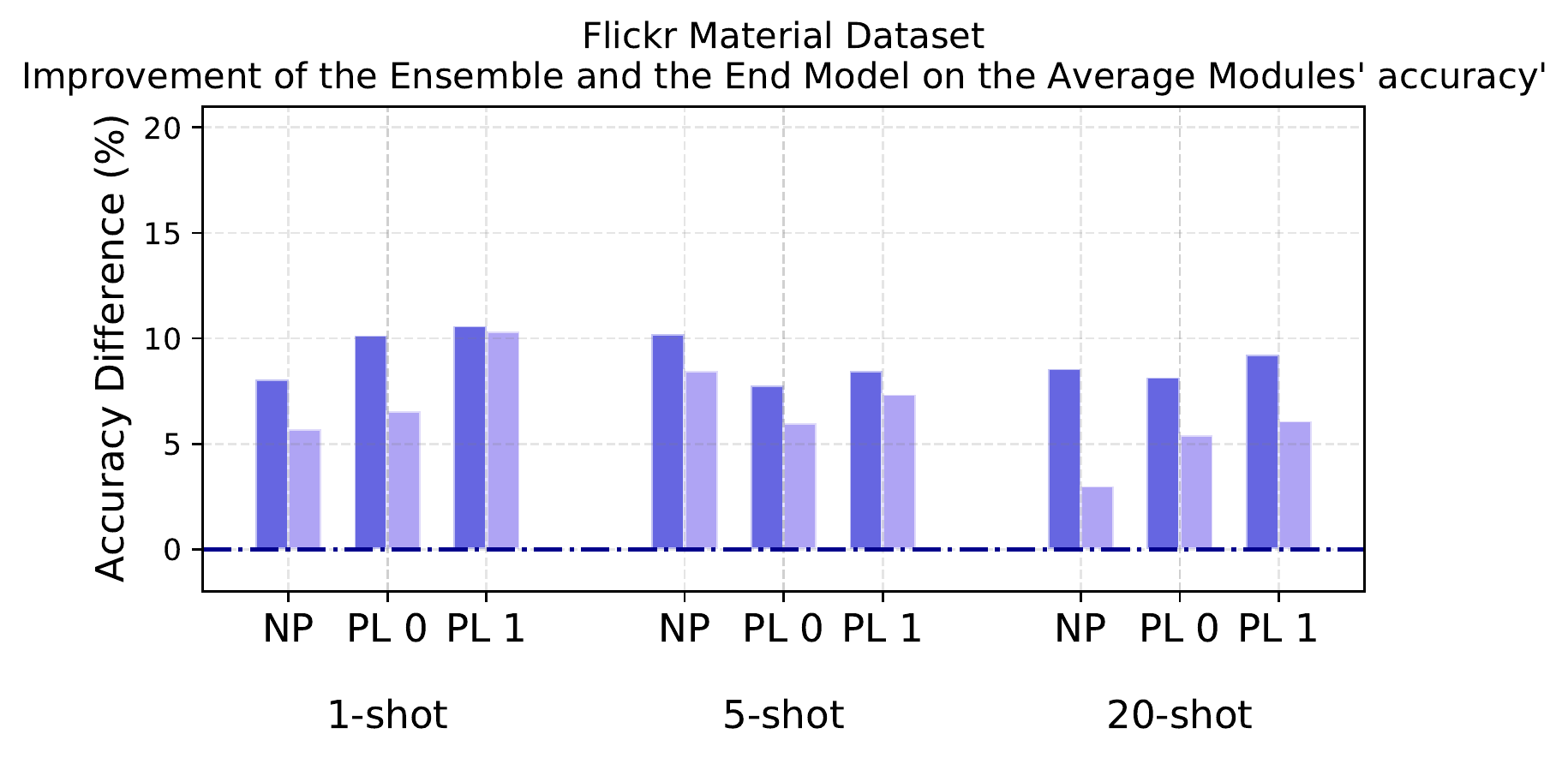}
\includegraphics[width=0.4\textwidth]{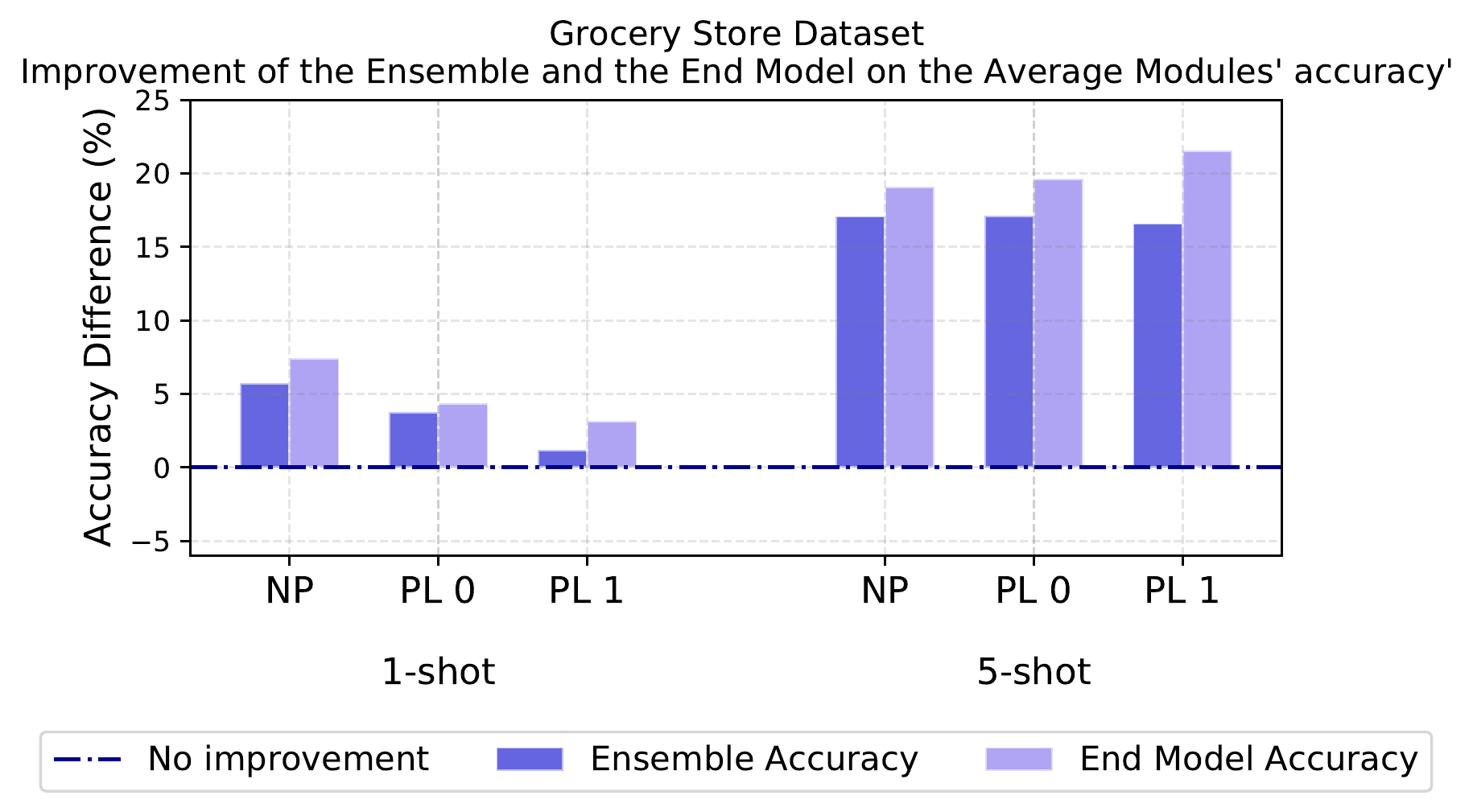}
\caption{(\texttt{Split 0}) For the \emph{OfficeHome-Clipart, Flickr Material}, and \emph{Grocery Store} datasets, we fix the number of labeled examples (1, 5, and 20 shots) and plot the improvements of the ensemble and the end model on the average accuracy of the training modules, for different pruning levels. The results refer to \sys with ResNet-50 backbone and the 95\% confidence intervals are computed over three different seeds.}
\label{fig:split_0_ensemble}
\end{figure}

In~\Cref{fig:split_0_ensemble}, we show that the ensemble generates improvements over the average modules accuracy in all the datasets. 
Consistently with findings in~\Cref{sec:intermediate_vs_final}, the relatedness of the auxiliary data to the target classes (i.e., levels of pruning) is not correlated to the effects of ensemble on the system accuracy.

\begin{figure}[t]
\centering
\includegraphics[width=0.4\textwidth]{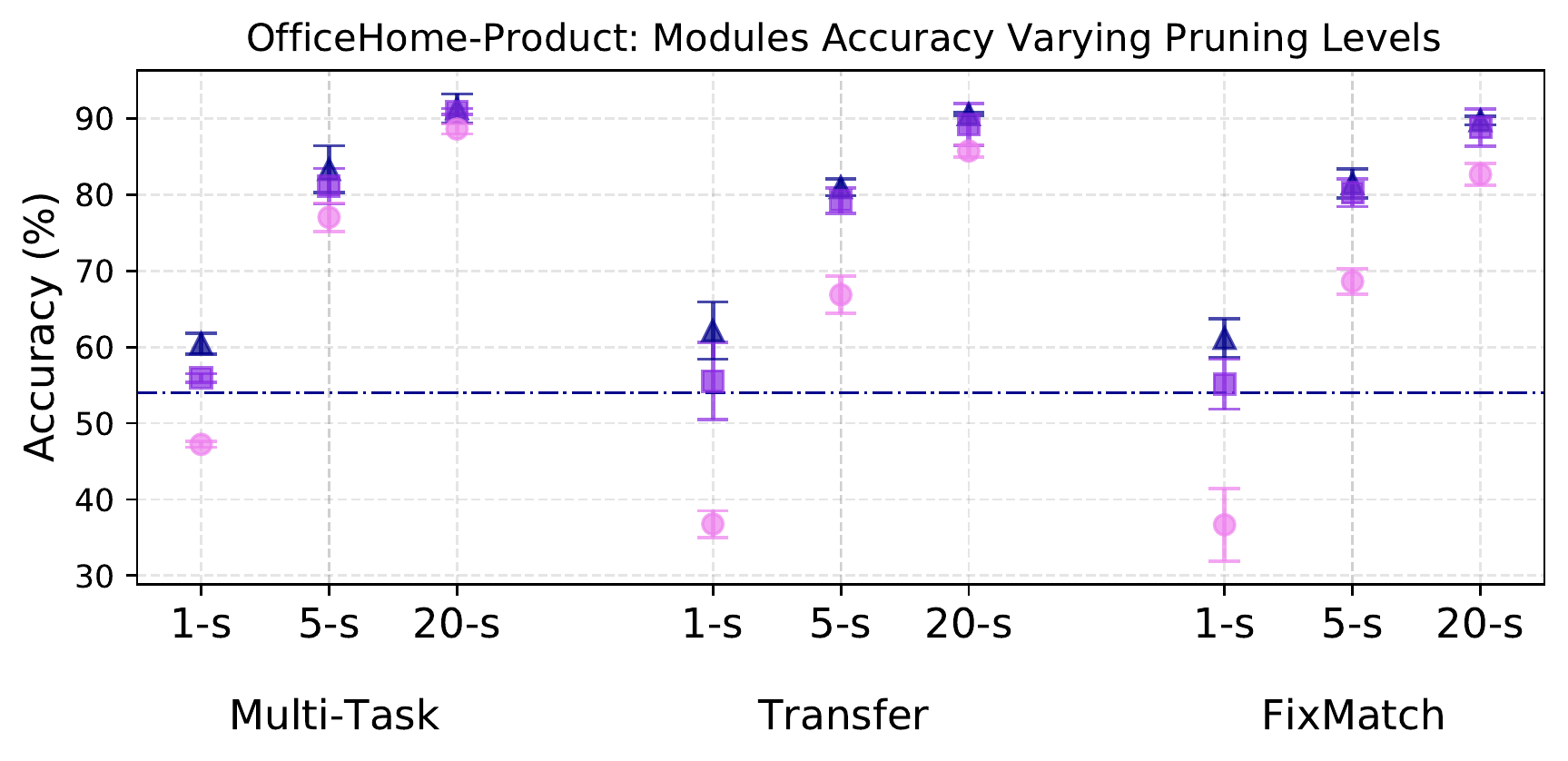}
\includegraphics[width=0.4\textwidth]{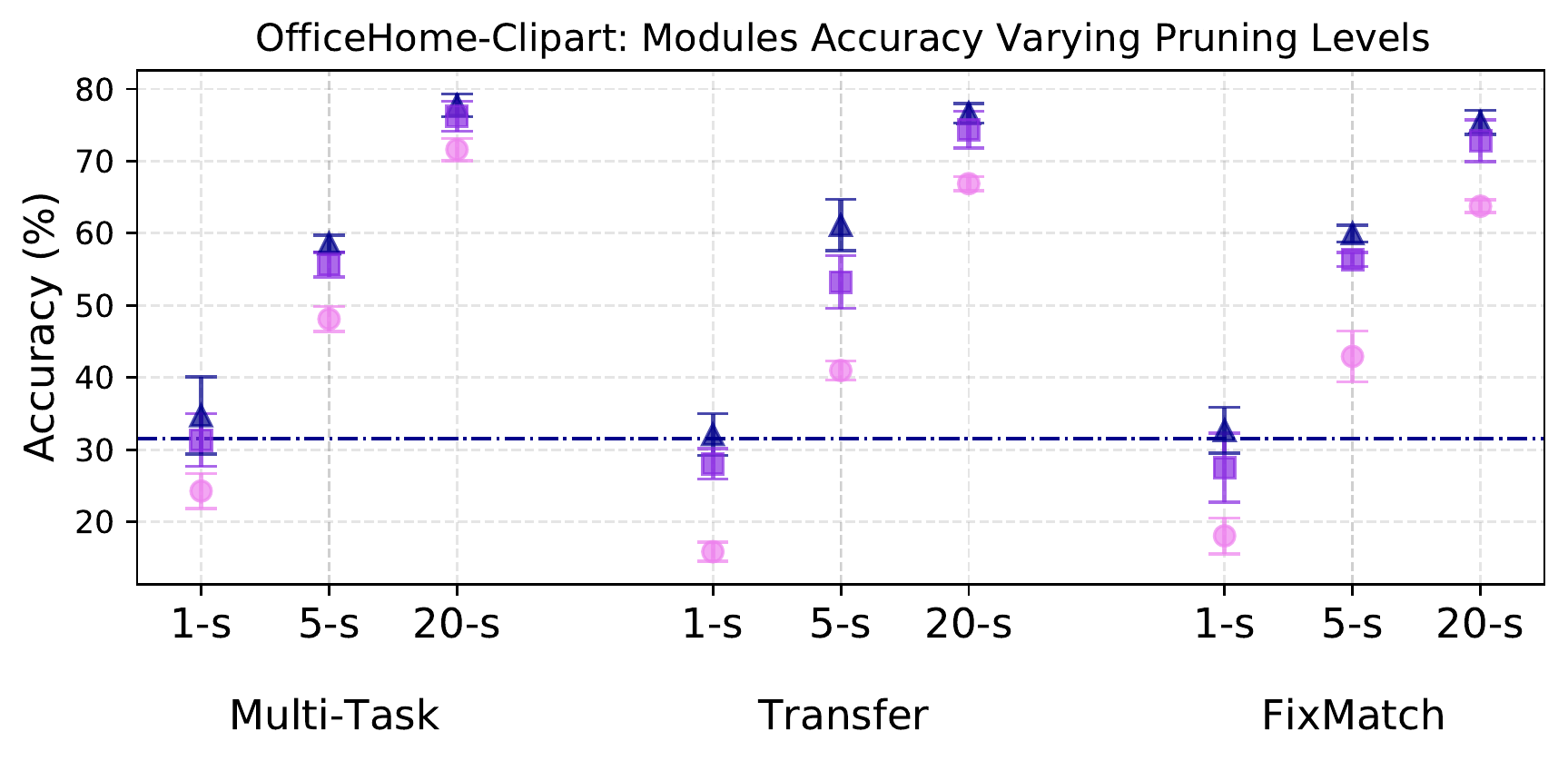}
\includegraphics[width=0.4\textwidth]{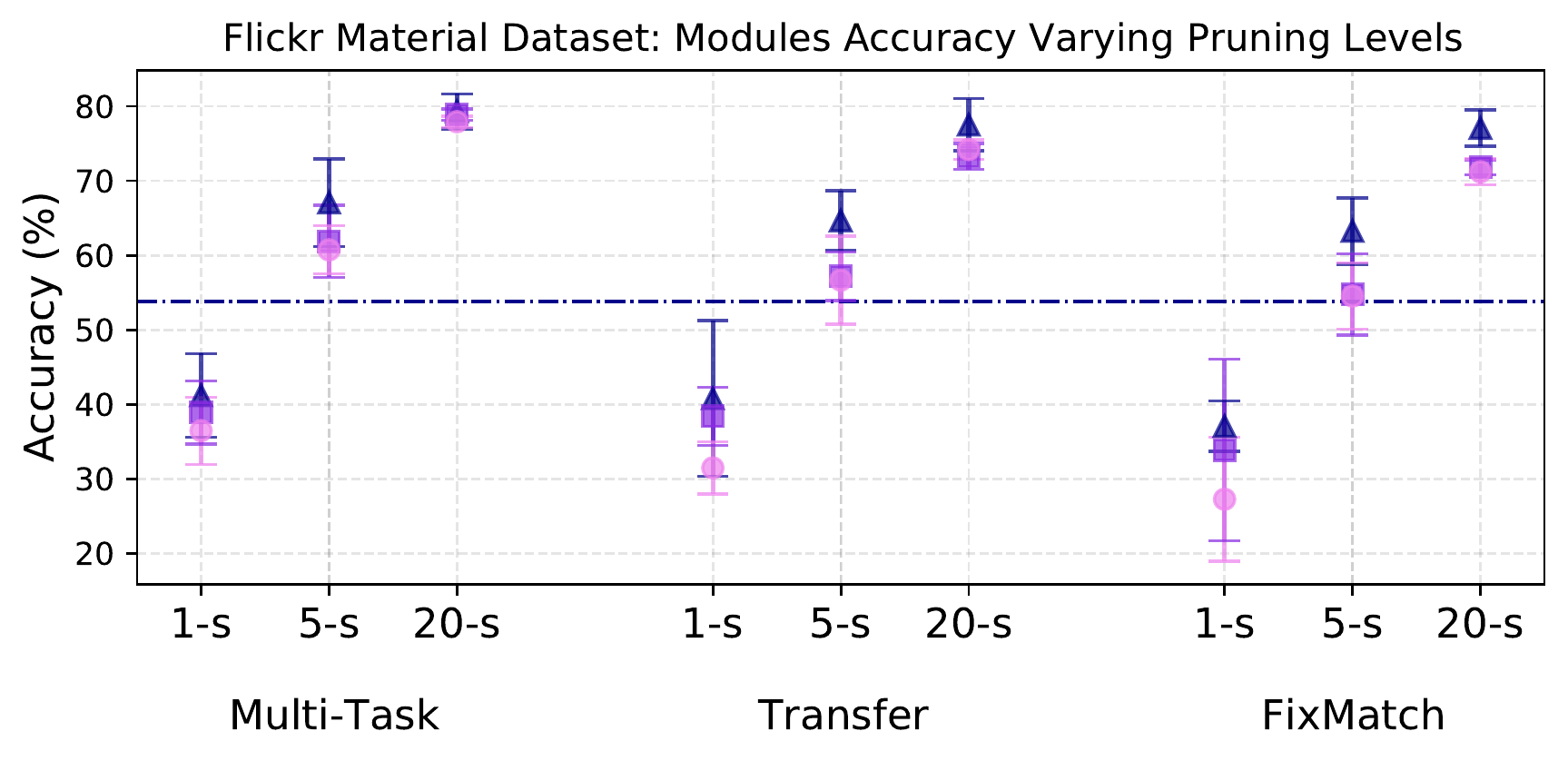}
\includegraphics[width=0.4\textwidth]{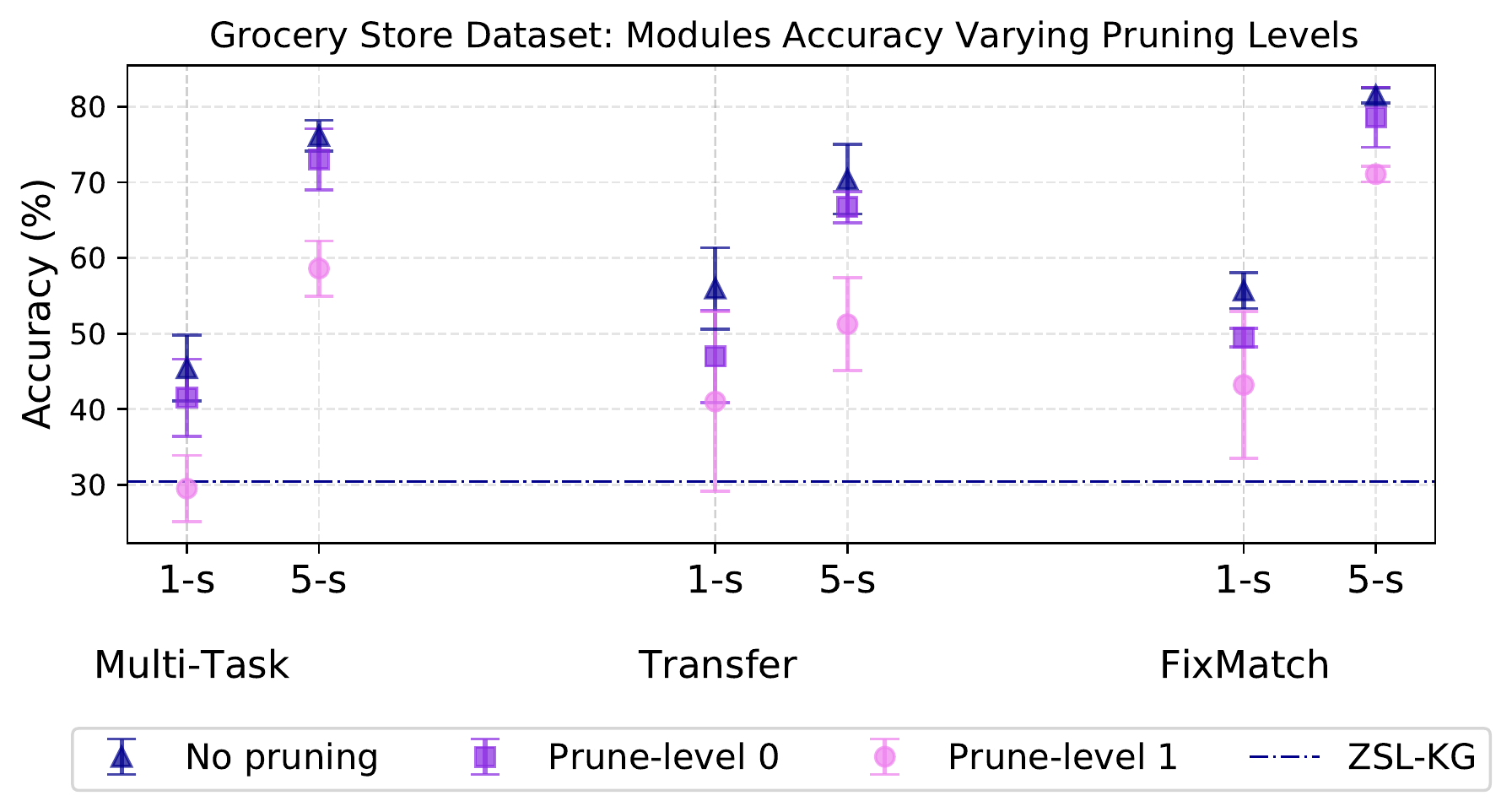}
\caption{(\texttt{Split 1}) The plot shows the \%-accuracy of each module at different levels of pruning (shapes) and for varying amount of labeled examples (x-axis), on the \emph{OfficeHome-Product}, \emph{OfficeHome-Clipart}, \emph{Flick Material} and \emph{Grocery Store} datasets.The error bars indicate 95\% confidence intervals on three seeds. All modules have ResNet-50 backbone. We note that the performances of ZSL-KG module are invariant to pruning, since it is used as pre-trained model on ConceptNet.}
\label{fig:split_1_pruning}
\end{figure}

\begin{figure}[tbp]
\centering
\includegraphics[width=0.4\textwidth]{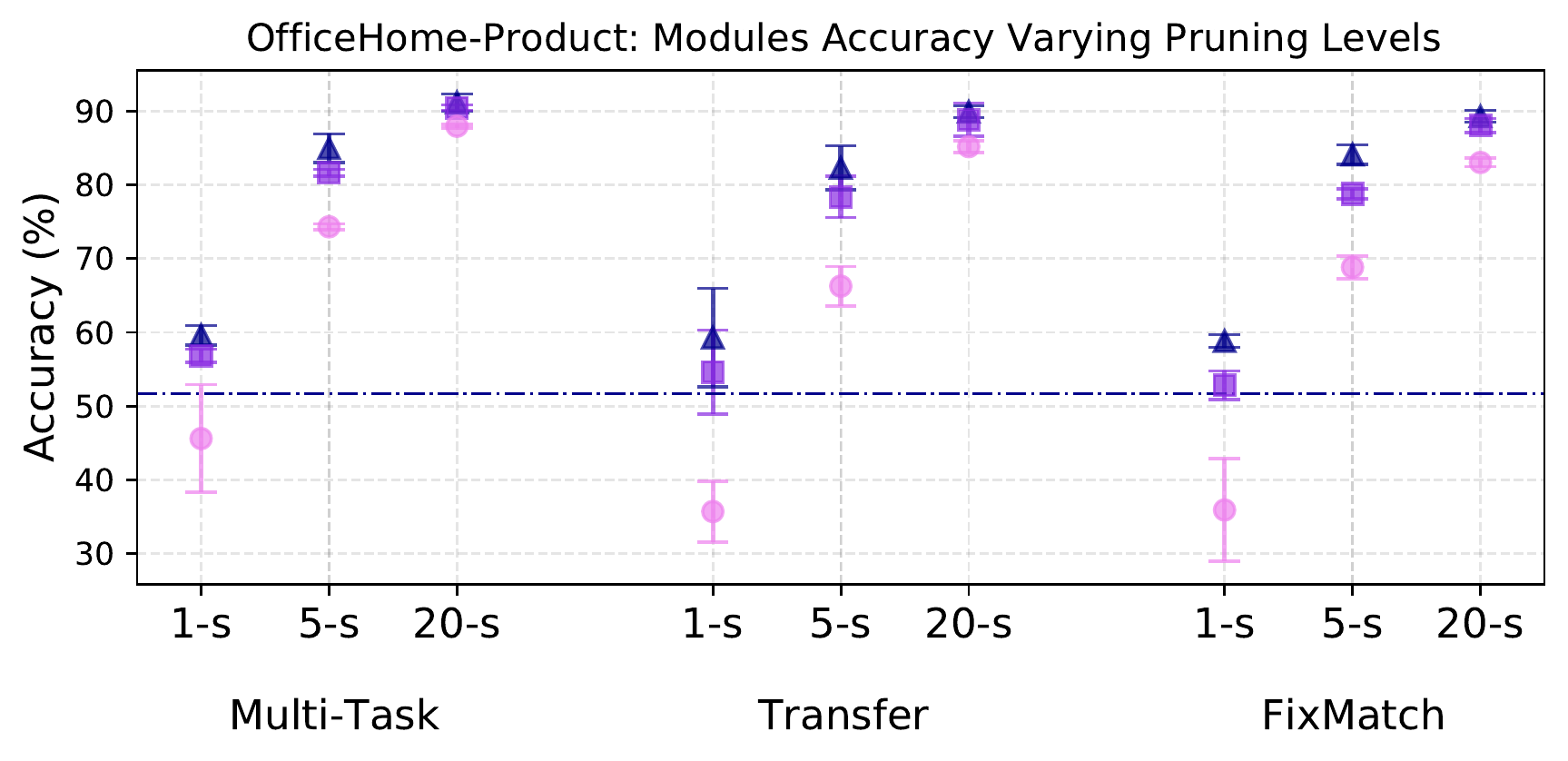}
\includegraphics[width=0.4\textwidth]{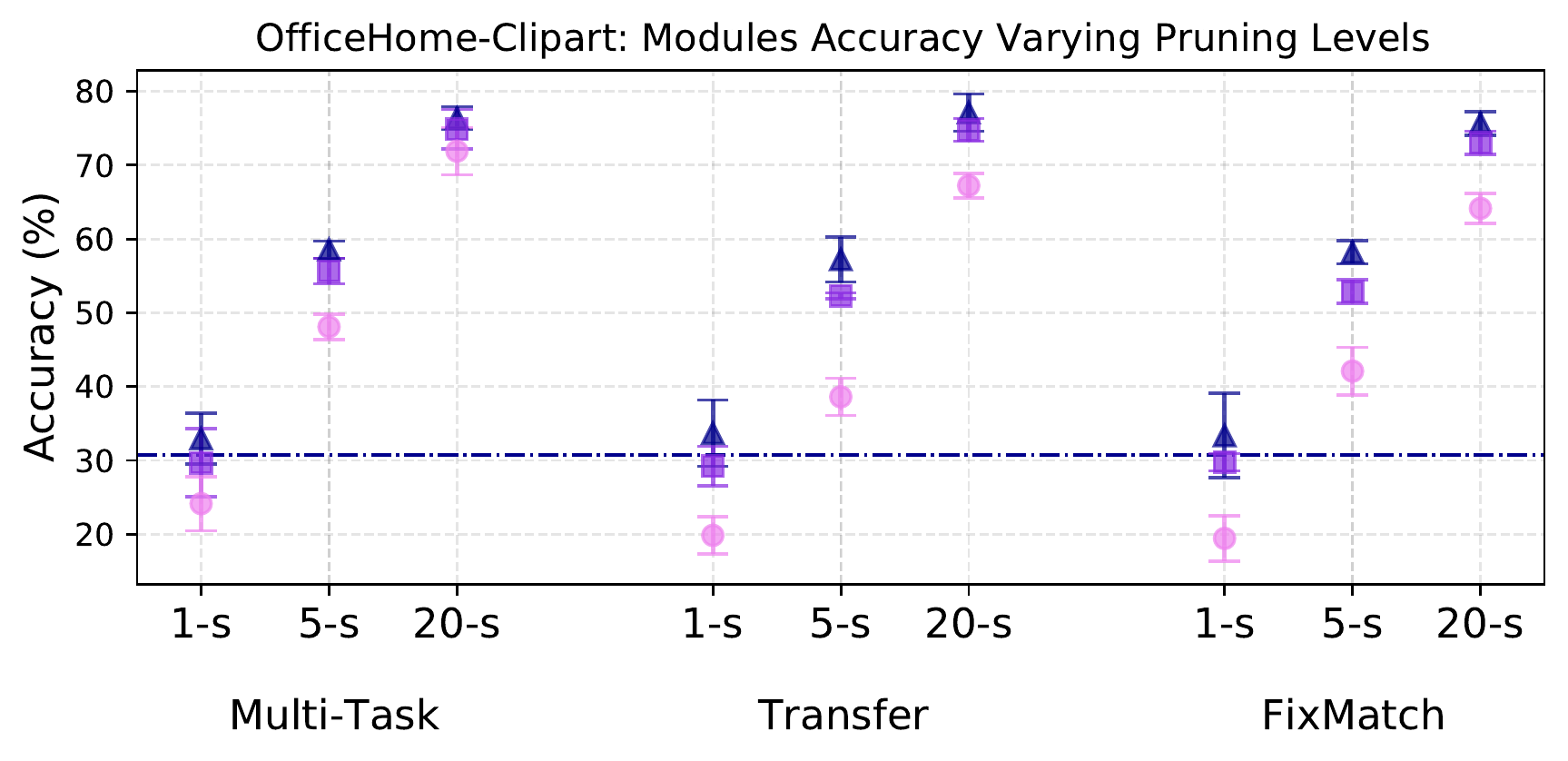}
\includegraphics[width=0.4\textwidth]{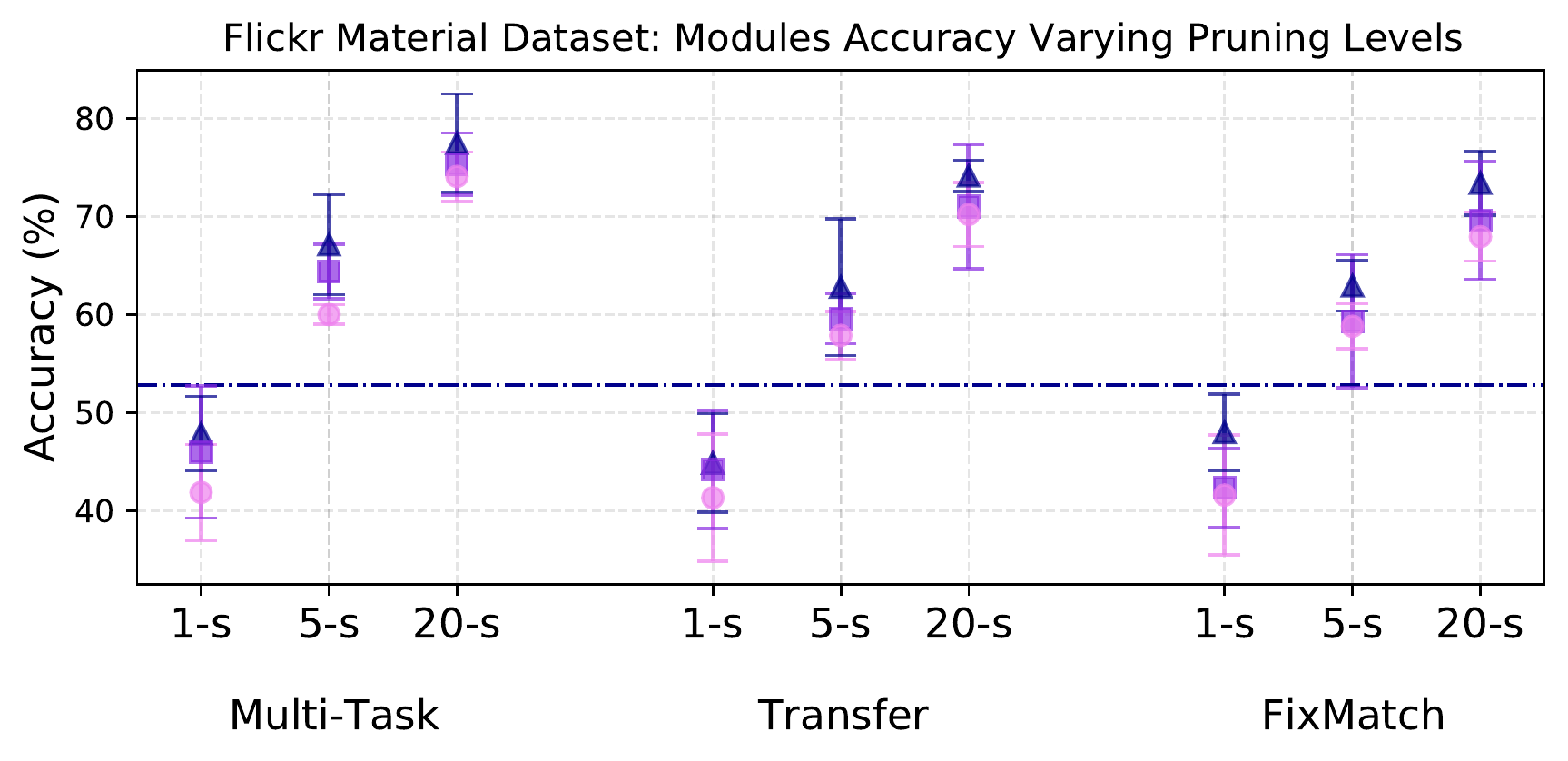}
\includegraphics[width=0.4\textwidth]{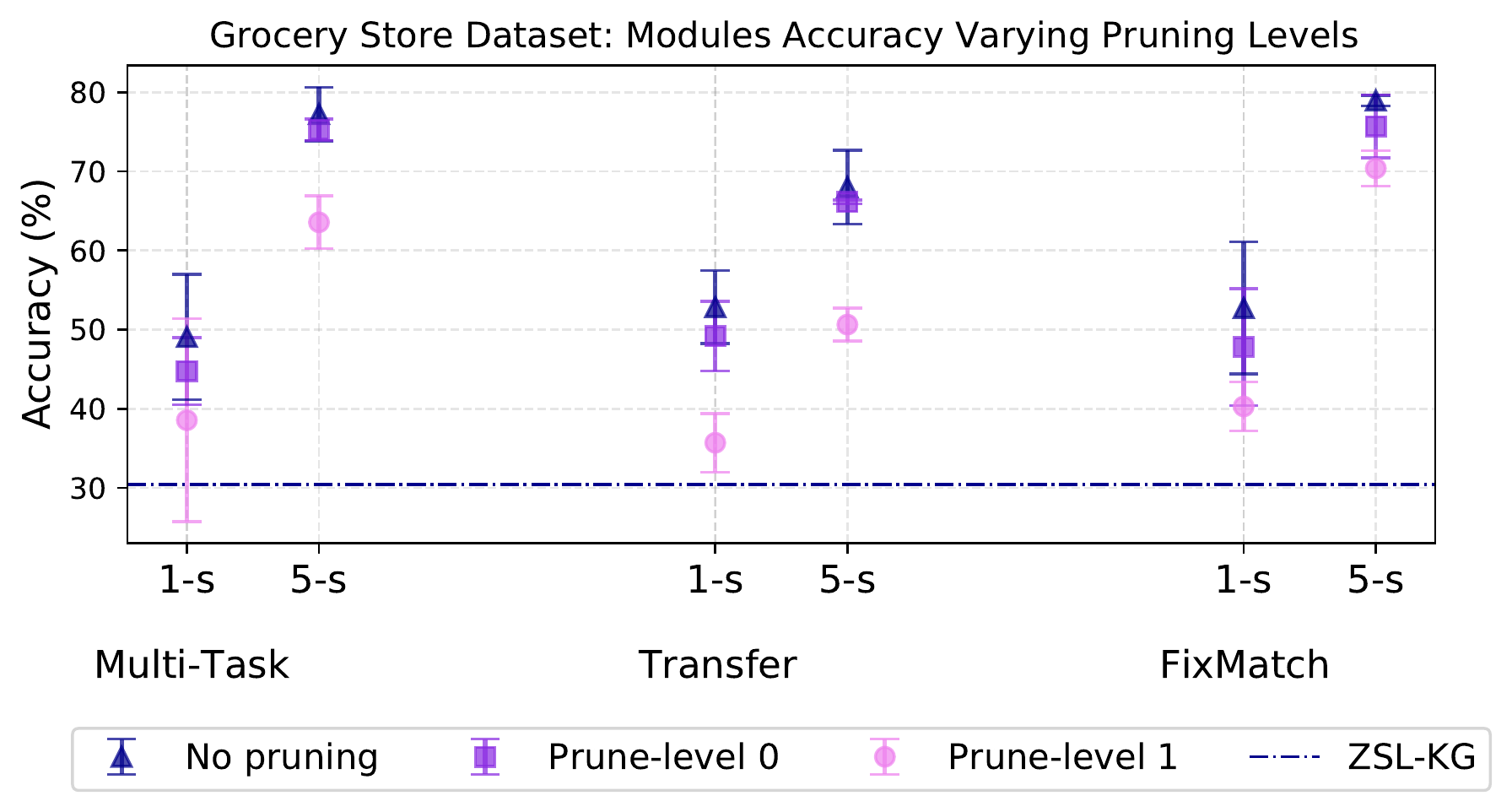}
\caption{(\texttt{Split 2}) The plot shows the \%-accuracy of each module at different levels of pruning (shapes) and for varying amount of labeled examples (x-axis), on the \emph{OfficeHome-Product}, \emph{OfficeHome-Clipart}, \emph{Flick Material} and \emph{Grocery Store} datasets.The error bars indicate 95\% confidence intervals on three seeds. All modules have ResNet-50 backbone. We note that the performances of ZSL-KG module are invariant to pruning, since it is used as pre-trained model on ConceptNet.}
\label{fig:split_2_pruning}
\end{figure}

\begin{figure}[ht]
\centering
\includegraphics[width=0.4\textwidth]{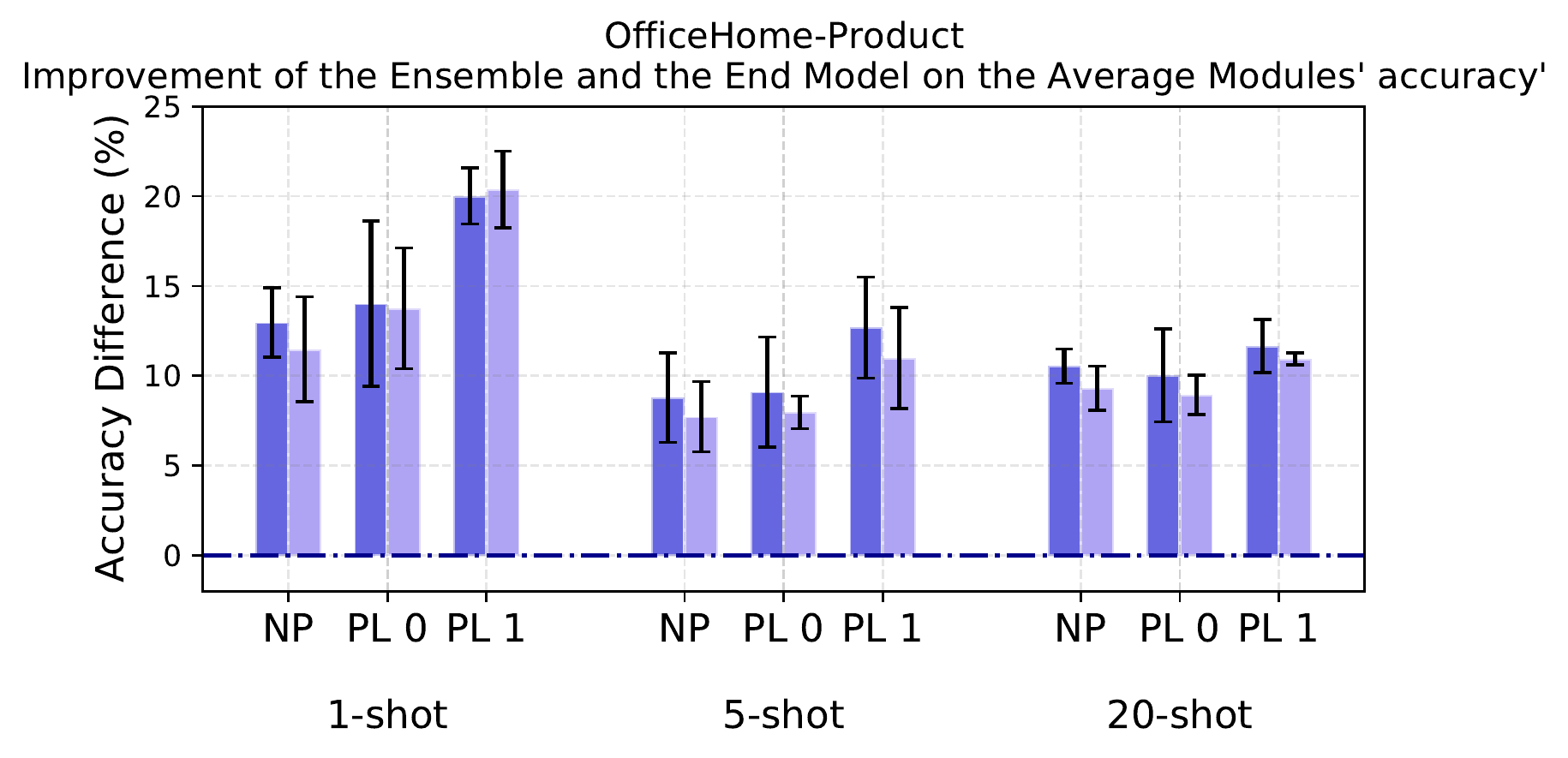}
\includegraphics[width=0.4\textwidth]{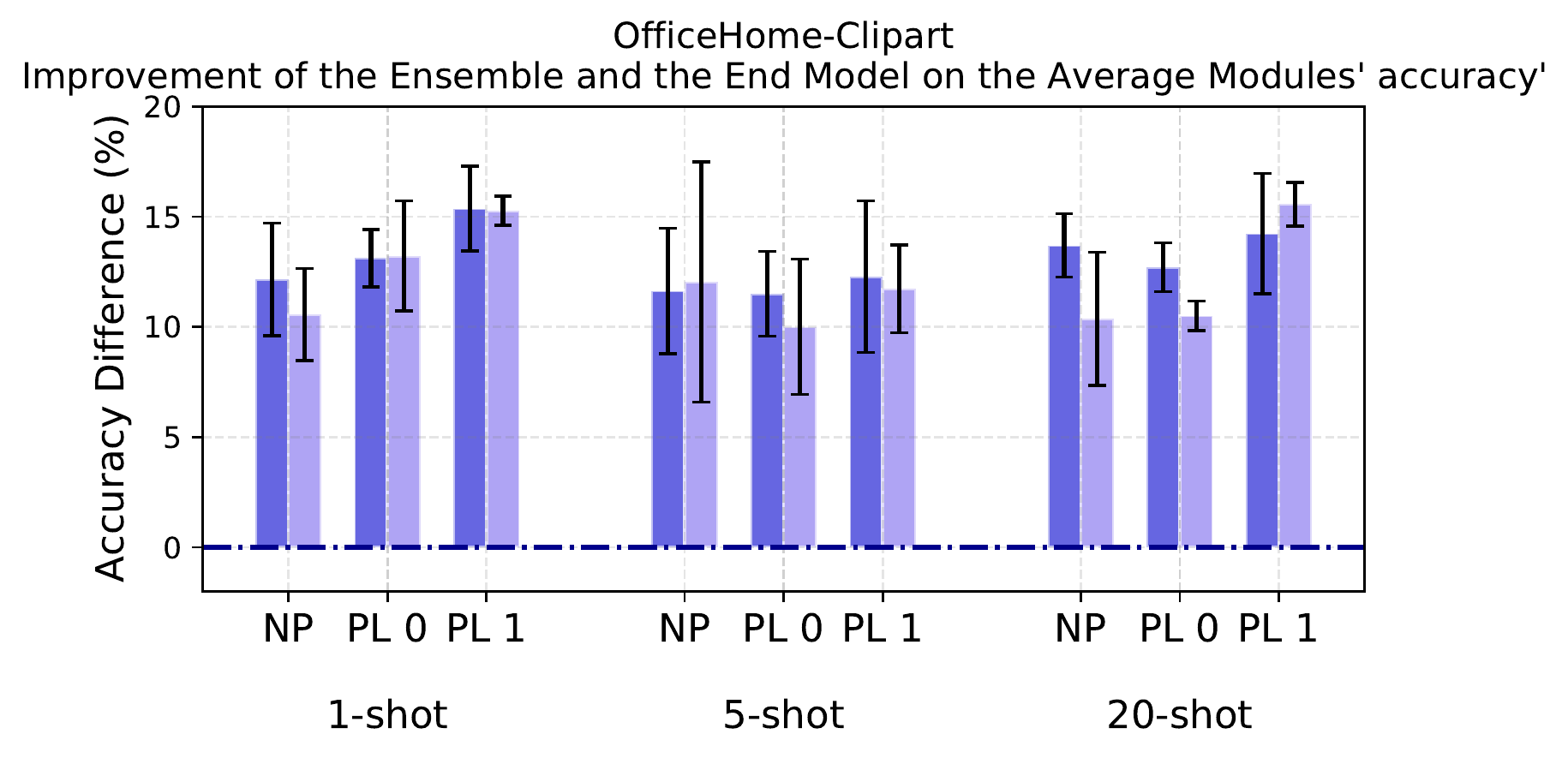}
\includegraphics[width=0.4\textwidth]{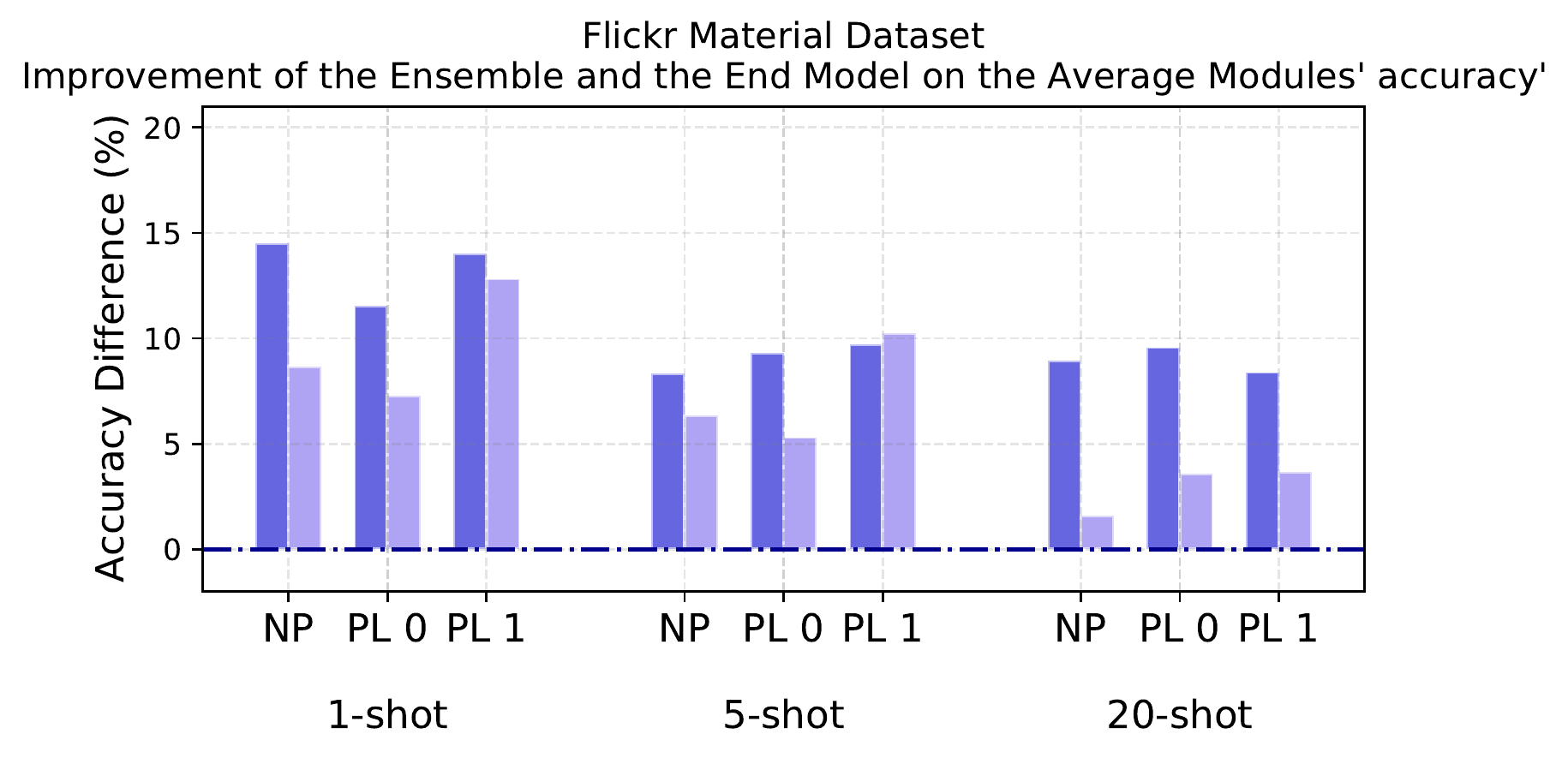}
\includegraphics[width=0.4\textwidth]{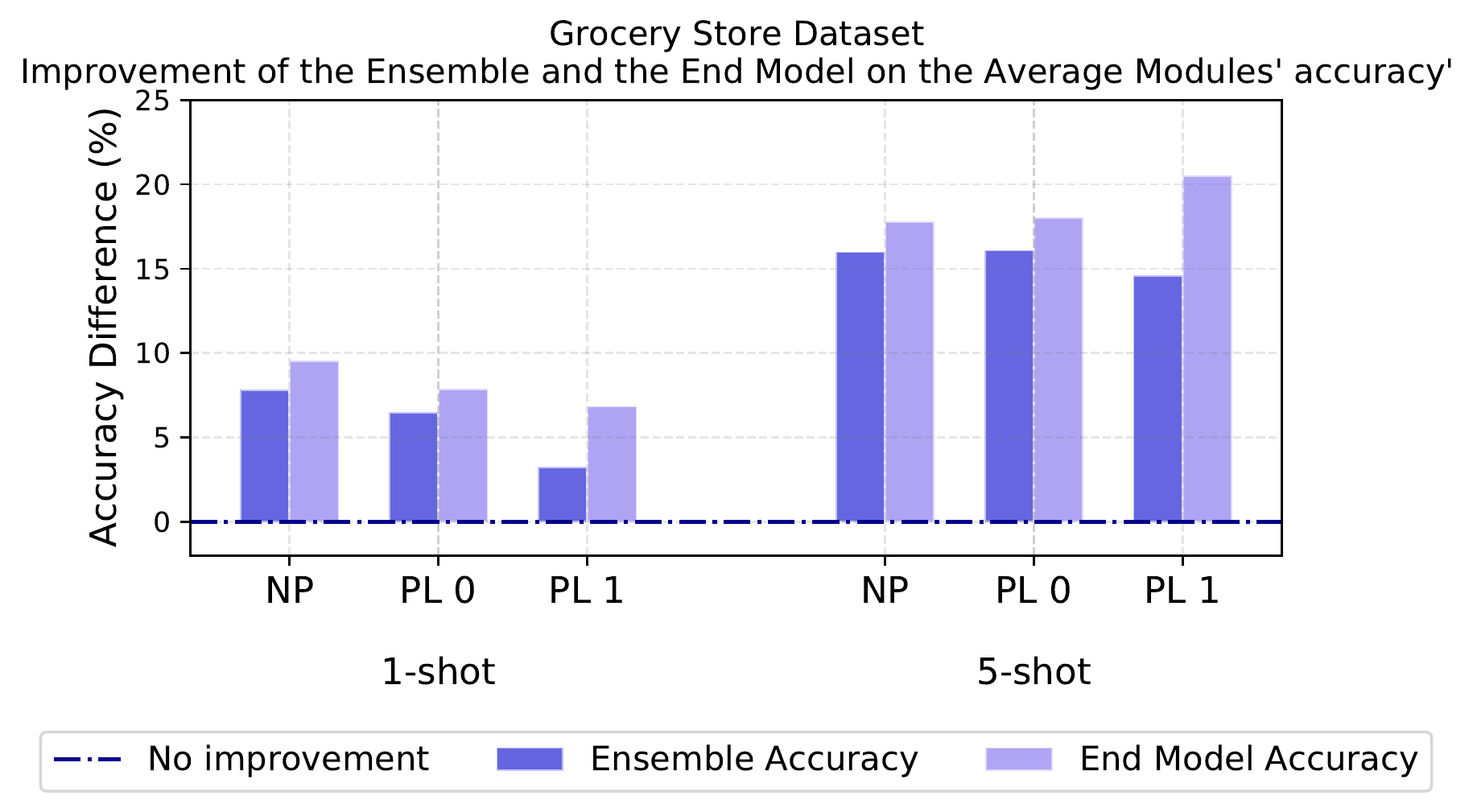}
\caption{(\texttt{Split 1}) For the \emph{OfficeHome-Product}, \emph{OfficeHome-Clipart, Flickr Material}, and \emph{Grocery Store} datasets, we fix the number of labeled examples (1, 5, and 20 shots) and plot the improvements of the ensemble and the end model on the average accuracy of the training modules, for different pruning levels. The results refer to \sys with ResNet-50 backbone and the 95\% confidence intervals are computed over three different seeds.}
\label{fig:split_1_ensemble}
\end{figure}

\begin{figure}[ht]
\centering
\includegraphics[width=0.4\textwidth]{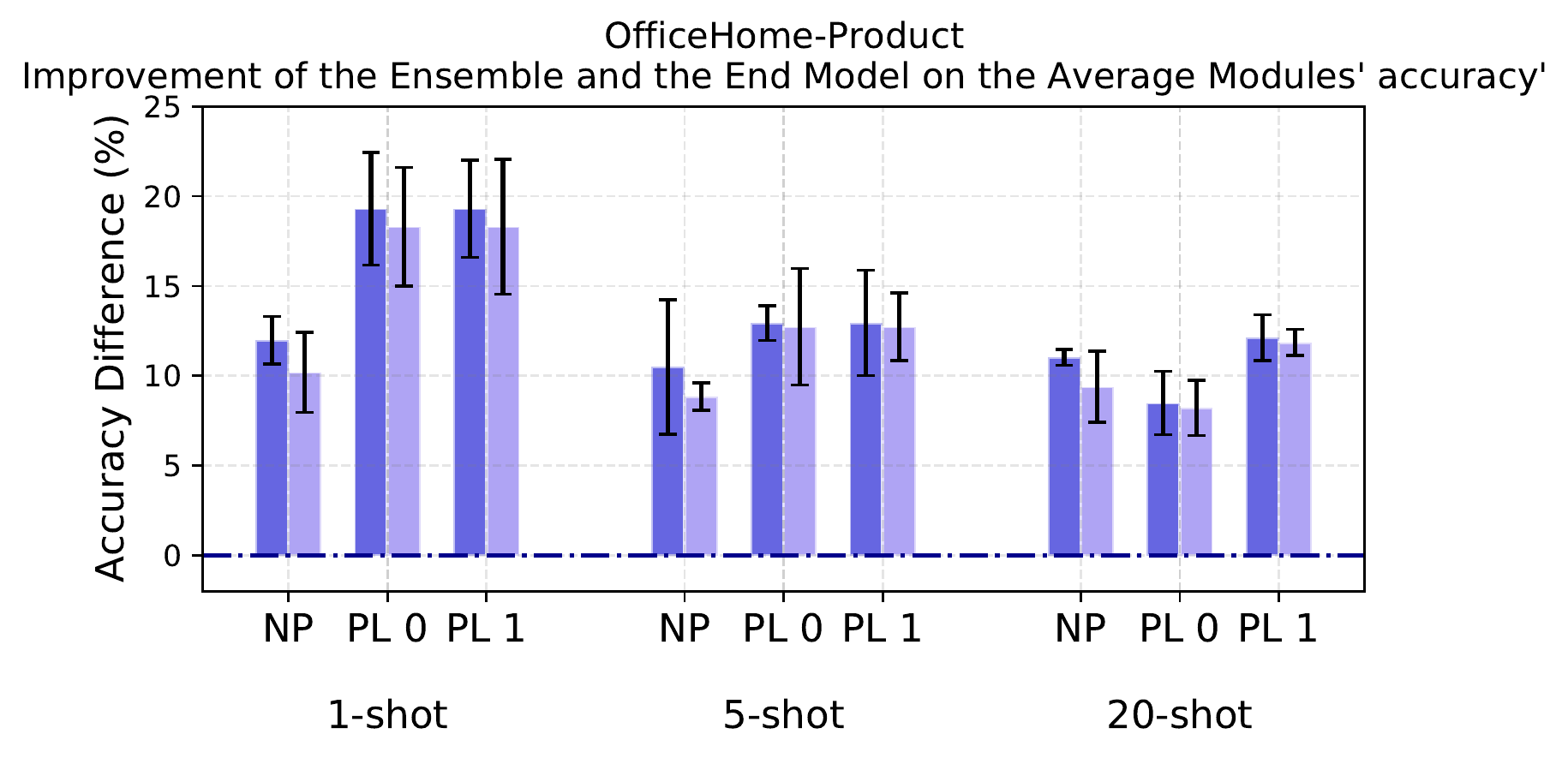}
\includegraphics[width=0.4\textwidth]{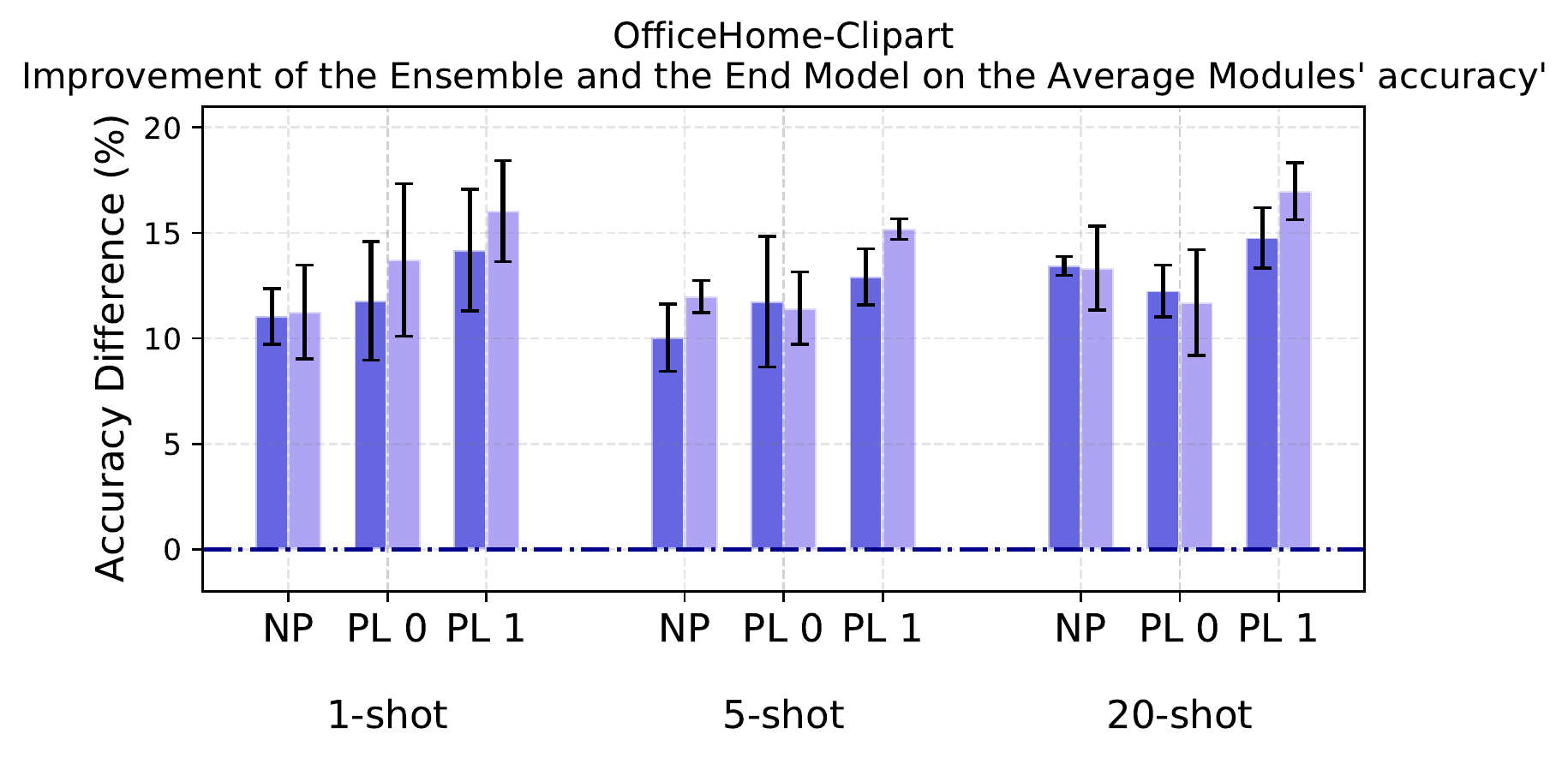}
\includegraphics[width=0.4\textwidth]{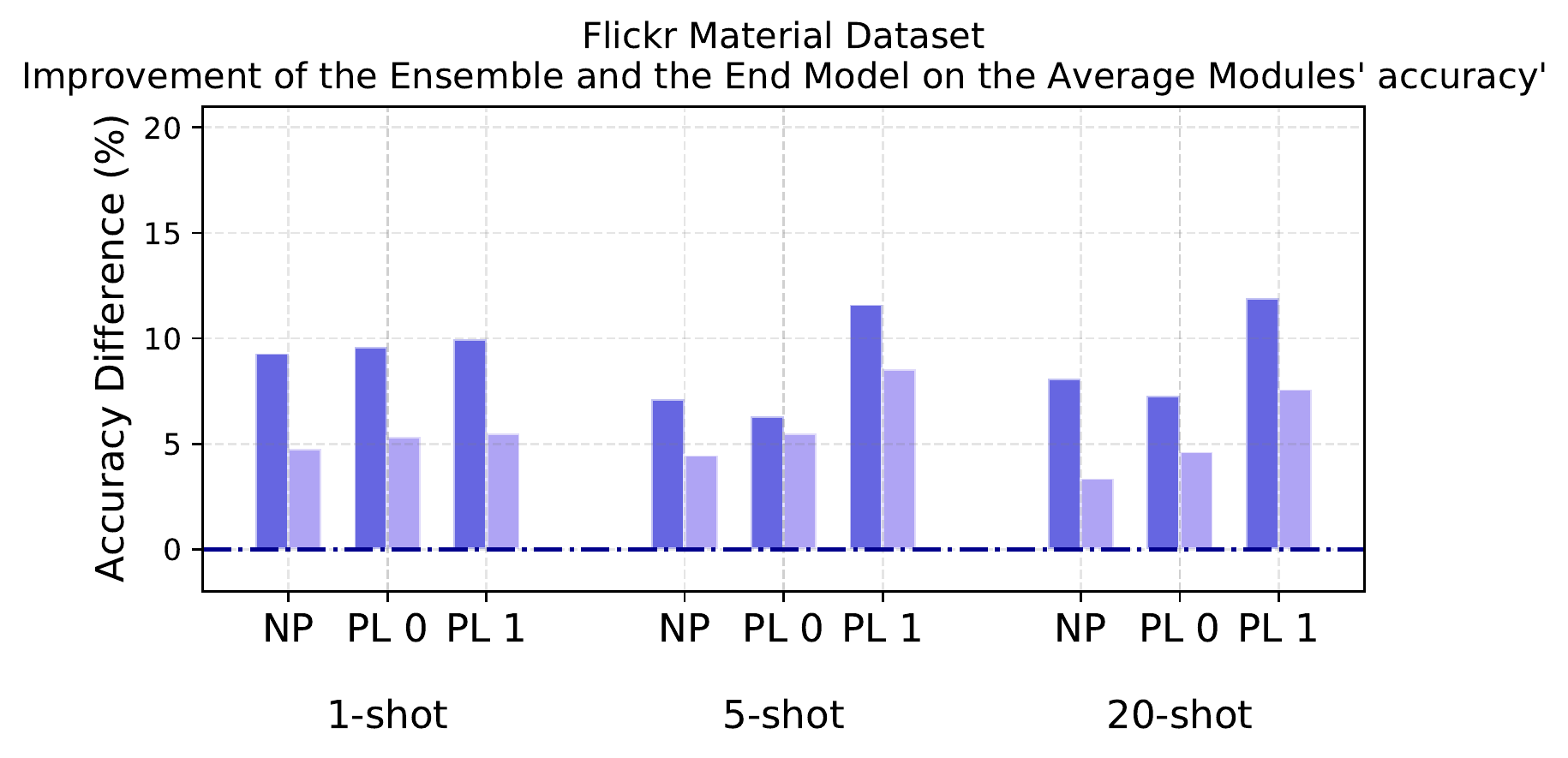}
\includegraphics[width=0.4\textwidth]{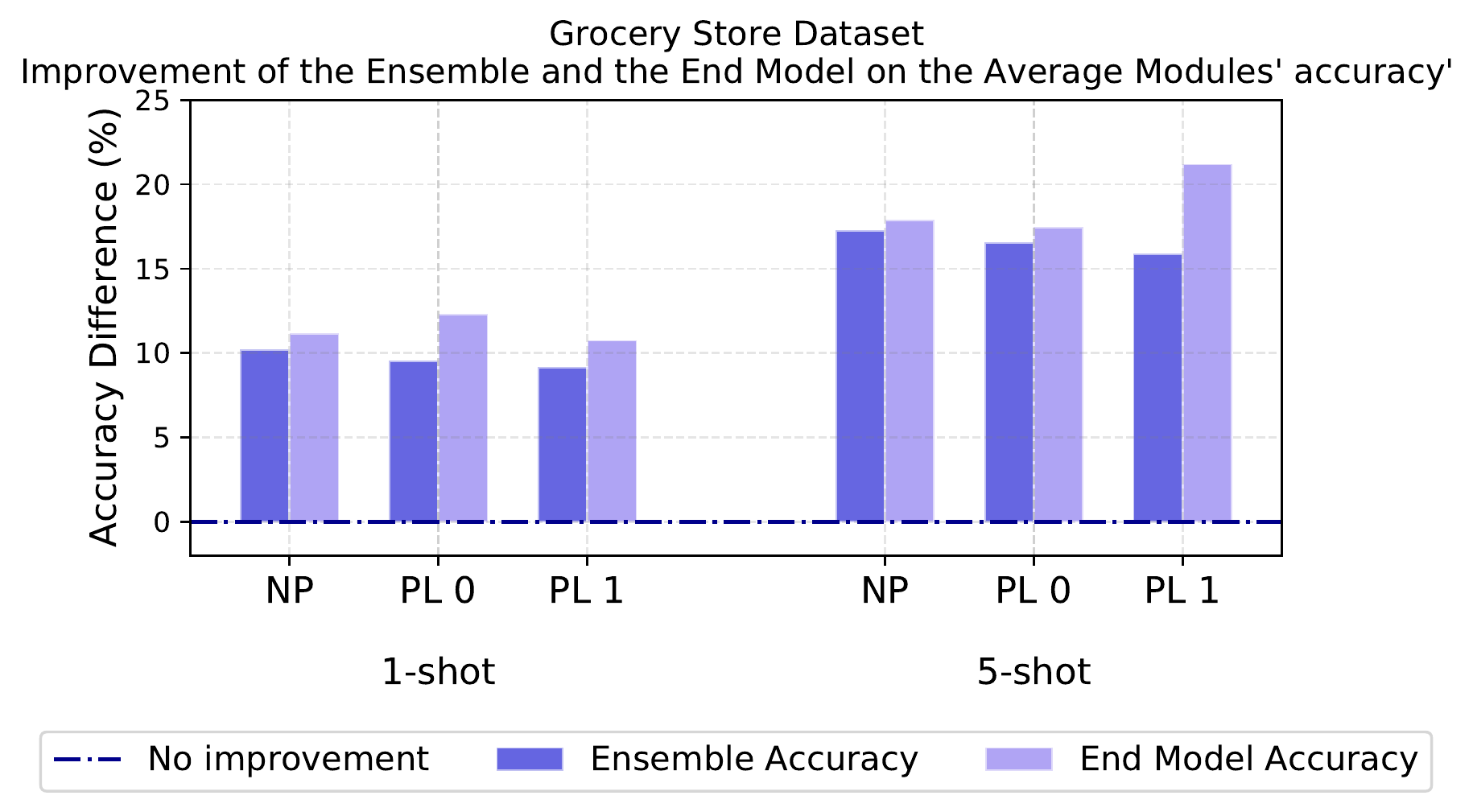}
\caption{(\texttt{Split 2}) For the \emph{OfficeHome-Product}, \emph{OfficeHome-Clipart, Flickr Material}, and \emph{Grocery Store} datasets, we fix the number of labeled examples (1, 5, and 20 shots) and plot the improvements of the ensemble and the end model on the average accuracy of the training modules, for different pruning levels. The results refer to \sys with ResNet-50 backbone and the 95\% confidence intervals are computed over three different seeds.}
\label{fig:split_2_ensemble}
\end{figure}

\section{Artifact Appendix}

\subsection{Abstract}

Our artifact provides the source code of TAGLETS, installation instructions, and the necessary scripts to run a demo. 
The target task of the demo is CIFAR-10 with limited labeled data, and we solve it accessing to CIFAR-100 as the auxiliary data. 
To replicate the results in the paper, readers must download the datasets used for the experiments, which we refer to in~\Cref{data_download}.

\subsection{Artifact check-list (meta-information)}

{\small
\begin{itemize}
  \item {\bf Algorithm: }Yes
  \item {\bf Data set: } CIFAR-10 (demo), CIFAR-100 (demo), OfficeHome (experiments), Grocery Store Dataset (experiments), Flickr Material Dataset (experiments), ImageNet21k (experiments)
  \item {\bf Run-time environment: } Debian GNU/Linux with (optional) CUDA 9 and CUDNN
  \item {\bf Hardware:} CPU or GPU (recommended)
  \item {\bf Metrics:} Top-1 accuracy
  \item {\bf Output:} Log files with accuracy
  \item {\bf Experiments:} We provide instructions to install the system and run a demo in the README file of the repository indicated in the following sections.
  \item {\bf How much disk space required (approximately)?} 15 Gb  
  \item {\bf How much time is needed to complete experiments (approximately)?} 2 hours with 1 NVIDIA V100 GPUs/45 minutes with 4 NVIDIA V100 GPUs (demo)
  \item {\bf Publicly available?} Yes
  \item {\bf Code licenses (if publicly available)?} Apache License 2.0
  \item {\bf Archived (provide DOI)? }\url{10.5281/zenodo.6080805}
\end{itemize}}

\subsection{Description}

\subsubsection{How delivered}

Two open-sourced versions:
\begin{itemize}
    \item \verb|taglets.zip| on Zenodo.
    
    \url{https://doi.org/10.5281/zenodo.6080805}
    \item \verb|BatsResearch/taglets|, public git repository, latest version.
    
    \url{https://github.com/BatsResearch/taglets}

\end{itemize}

\subsubsection{Hardware dependencies}

The CPUs or GPUs in use should support Python 3.7 and Pytorch 1.7.

\subsubsection{Software dependencies}\label{app:software}

\begin{itemize}
    \item Python (3.7)
    \item PyTorch (1.7)
\end{itemize}

\subsubsection{Data sets}\label{data_download}

In the demo, we use CIFAR-10 and CIFAR-100 datasets. 
The GitHub repository includes the bash scripts to download them.
You can download the datasets used in the paper here:
\begin{itemize}
    \item \emph{OfficeHome}: \url{https://www.hemanthdv.org/officeHomeDataset.html}
    \item \emph{Grocery Store Dataset}: \url{https://github.com/marcusklasson/GroceryStoreDataset}
    \item \emph{Flickr Material Dataset}: \url{https://people.csail.mit.edu/lavanya/fmd.html}
\end{itemize}

\subsection{Installation}
\sys is implemented in Python.
Once the software requirements are satisfied (\Cref{app:software}), you can run the following script to install the system.
We note that these instruction are also reported in the GitHub README file.

Clone this repository:
\begin{small}
\begin{verbatim}
git clone 

https://github.com/BatsResearch/taglets.git
\end{verbatim}
\end{small}

Before installing TAGLETS, we recommend creating and activating a new virtual environment which can be done by the following script:
\begin{verbatim}
python -m venv taglets_venv
source taglets_venv/bin/activate
\end{verbatim}

You also want to make sure `pip` is up-to-date:
\begin{verbatim}
pip install --upgrade pip
\end{verbatim}

Then, to install \sys and download related files, run:
\begin{verbatim}
bash setup.sh
\end{verbatim}

\subsection{Experiment workflow}

We invite the reader to follow the workflow presented in the GitHub README file to run the demo. We recommend running it with GPUs to speed up the training.

\subsection{Evaluation and expected result}

For the demo, please follow the instructions on the README to run it. The fine-tuning baseline for the demo task is 41.5\% accuracy. The demo is expected to significantly outperforms this baseline.

To replicate results on this paper, you first have to download the datasets used in the paper, including OfficeHome, Grocery Store, Flickr Material, and ImageNet-21k.
We have provided the script to download the SCADS file used in the paper (installed with ImageNet-21k).
Then, set the hyperparameters to be the same as the ones used in this paper. 
Finally, run \sys with all four modules used in this paper.
You can expect the same results as presented in~\Cref{tab:acc:gs_and_fms,tab:acc:officehome} of the main body of the paper.

\subsection{Experiment customization}\label{exp:custom}

\sys can be used as a Python package where you can import it and interact with its interface. The two key classes of \sys that users will interact with are \verb|Controller| and \verb|Task|.

In \verb|Task|, there are the following argumentts:
\begin{itemize}
    \item \verb|input_shape| (required): image size
    \item \verb|batch_size| (optional, default to 128): batch size
    \item \verb|wanted_num_related_class| (optional, default to 10): number of auxiliary classes per target class
\end{itemize}
You can also choose a backbone of your choice to be used in \sys by using the \verb|set_initial_model| and \verb|set_model_type| methods of the \verb|Task| class.

In \verb|Controller|, you can specify the modules you want to use in \sys by setting the parameter \verb|modules|.

You can refer to the demo on how these parameters can be specified.
\subsection{Notes}

For more questions, please file issues on GitHub.


\end{document}